\documentclass[journal]{IEEEtran}

\usepackage{epsfig} 
\usepackage{mathptmx} 
\usepackage{times} 
\usepackage{amsmath} 
\usepackage{amssymb} 
\usepackage{amsthm}
\usepackage[table]{xcolor}
\usepackage{booktabs}
\usepackage{makecell}
\usepackage{multirow}
\usepackage[T1]{fontenc}

\usepackage{multirow}
\usepackage{float}
\usepackage{bm}
\usepackage{cite}
\usepackage{hyperref}
\usepackage{algorithm}
\usepackage{algorithmic}
\usepackage{enumitem}
\usepackage{subcaption}
\usepackage{makecell}

\definecolor{custompink}{RGB}{239,118,186} 
\hypersetup{
    colorlinks=true, 
    linkcolor=blue,  
    citecolor=green, 
    filecolor=magenta, 
    urlcolor=custompink     
}

\title{PierGuard: A Planning Framework for Underwater Robotic Inspection of Coastal Piers}

\author{Pengyu Wang$^{1, 2}$, Hin Wang Lin$^{2}$, Jialu Li$^{2}$, Jiankun Wang$^{1, 3}$, \textit{Senior Member, IEEE}, \\ Ling Shi$^{2}$, \textit{Fellow, IEEE} and Max Q.-H. Meng$^{1, 4}$, \textit{Fellow, IEEE}
\thanks{$^{1}$Pengyu Wang, Jiankun Wang and Max Q.-H. Meng are with the Shenzhen Key Laboratory of Robotics Perception and Intelligence and the Department of Electronic and Electrical Engineering, Southern University of Science and Technology, Shenzhen, China. {\tt\small pwangat@connect.ust.hk, wangjk@sustech.edu.cn, max.meng@ieee.org}}
\thanks{$^{2}$Pengyu Wang, Hin Wang Lin, Jialu Li and Ling Shi are with the Department of Electronic and Computer Engineering, Hong Kong University of Science and Technology, Hong Kong SAR. {\tt\small \{pwangat, hwlinaa, jlikr\}@connect.ust.hk, eesling@ust.hk}}%
\thanks{$^{3}$Jiankun Wang is also with the Jiaxing Research Institute, Southern University of Science and Technology, Jiaxing, China.}
\thanks{$^{4}$Max Q.-H. Meng is also a Professor Emeritus in the Department of Electronic Engineering at The Chinese University of Hong Kong in Hong Kong and was a Professor in the Department of Electrical and Computer Engineering at the University of Alberta in Canada.}
}

\begin{document}

\maketitle

\begin{abstract}
Using underwater robots instead of humans for the inspection of coastal piers can enhance efficiency while reducing risks. A key challenge in performing these tasks lies in achieving efficient and rapid path planning within complex environments. Sampling-based path planning methods, such as Rapidly-exploring Random Tree* (RRT*), have demonstrated notable performance in high-dimensional spaces. In recent years, researchers have begun designing various geometry-inspired heuristics and neural network-driven heuristics to further enhance the effectiveness of RRT*. However, the performance of these general path planning methods still requires improvement when applied to highly cluttered underwater environments. In this paper, we propose PierGuard, which combines the strengths of bidirectional search and neural network-driven heuristic regions. We design a specialized neural network to generate high-quality heuristic regions in cluttered maps, thereby improving the performance of the path planning. Through extensive simulation and real-world ocean field experiments, we demonstrate the effectiveness and efficiency of our proposed method compared with previous research. \textcolor{blue}{Our method achieves approximately 2.6 times the performance of the state-of-the-art geometric-based sampling method and nearly 4.9 times that of the state-of-the-art learning-based sampling method.} Our results provide valuable insights for the automation of pier inspection and the enhancement of maritime safety. The updated experimental video is available in the supplementary materials.
\end{abstract}

\def\abstractname{Note to Practitioners}
\begin{abstract}
The research presented in this paper focuses on enhancing the efficiency and safety of underwater inspections of coastal piers by utilizing underwater robots instead of human divers. One of the primary challenges in this context is the development of efficient path planning strategies within complex, dynamic environments. Our study investigates the use of sampling-based path planning methods, specifically RRT*, to improve the efficiency of these inspections. The key takeaway for practitioners is that integrating bidirectional search with neural heuristic regions in the path planning of underwater robots can significantly enhance inspection efficiency, reducing both the time required and the complexity of navigating hazardous environments. The method is particularly useful in environments where the terrain is difficult to model or constantly changing. By adopting these algorithms, practitioners can improve operational efficiency, enhance safety protocols, and reduce the overall cost of inspections. We recommend that practitioners looking to implement this approach invest in suitable robotic platforms capable of handling the complexities of underwater navigation. Additionally, integrating these path planning methods with real-time data acquisition systems will further enhance the effectiveness of inspections, allowing for more accurate and faster assessments of coastal infrastructure.
\end{abstract}

\def\abstractname{Index Terms}
\begin{abstract}
Neural network, path planning, sampling-based algorithm, underwater vehicle.
\end{abstract}

\begin{figure}[t!]
    \centering
    \begin{subfigure}[b]{0.45\textwidth}
        \includegraphics[width=\textwidth]{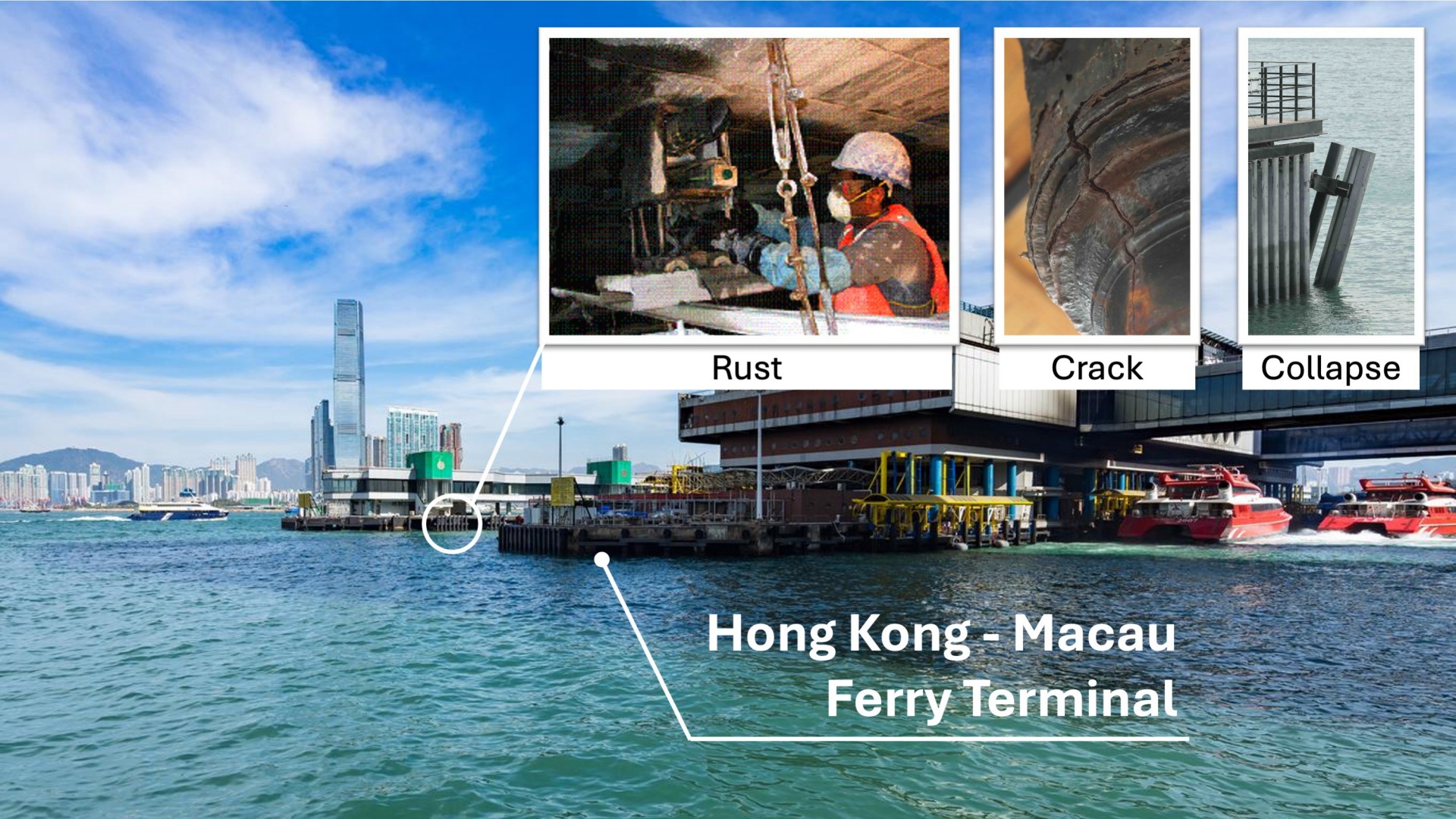} 
        \caption{Hong Kong - Macau Ferry Terminal and its inspection.}
        \label{fig:fig1_1}
    \end{subfigure}
    \begin{subfigure}[b]{0.45\textwidth}
        \includegraphics[width=\textwidth]{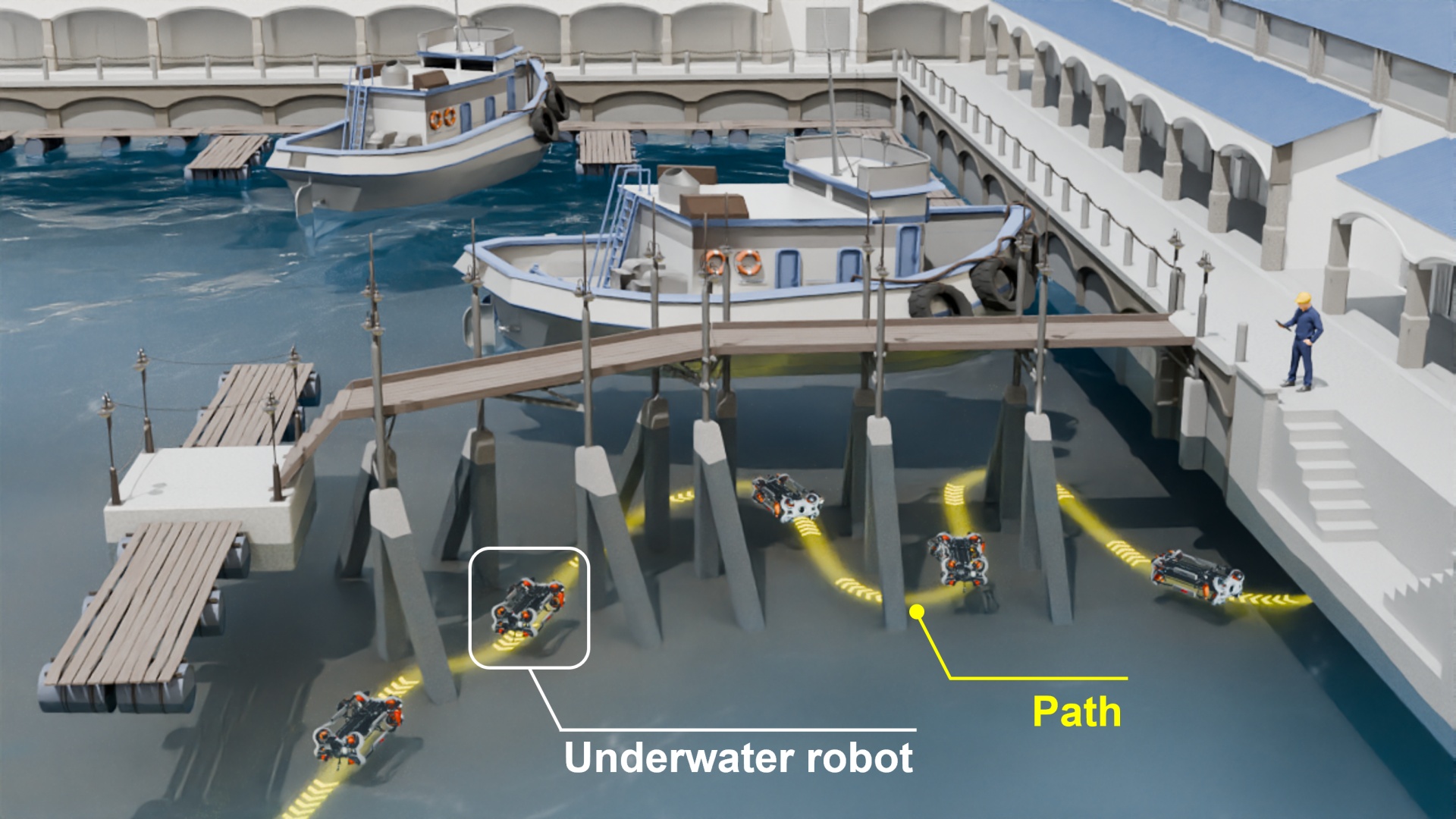} 
        \caption{Underwater robot inspects pier along the planned path.}
        \label{fig:fig1_2}
    \end{subfigure}
    \caption{Illustration of pier inspection: manual vs. robotic.}
    \label{fig:fig1}
\end{figure}

\section{INTRODUCTION}
\IEEEPARstart{I}{n} \textcolor{blue}{recent years, various forms of mobile robots, including ground~\cite{wendemagegn2024enhancing, kedir2024pso, wang2025mac}, aerial~\cite{wang2022quadrotor, gedefaw2024improved, madebo2024enhanced, abera2024improved}, and underwater~\cite{kim2024development, lin2024coastal} types, have achieved significant success in enhancing efficiency and reducing risks for humans.} In this paper, we focus on utilizing underwater robotic systems as an alternative to conventional divers for inspecting complex environments at pier bottoms. As two of the world's top ten ports, Hong Kong and Shenzhen have approximately 173 and 154 formal coastal piers of various sizes, respectively. During peak hours, ships dock at these piers every 15 minutes. Due to factors such as heavy operational activity and seawater erosion, the reliability and safety of pier structures have become critically important. As illustrated in Fig.~\ref{fig:fig1_1}\footnote{Source: Adapted from CEDD of Hong Kong and HK01.}, the Hong Kong-Macau Ferry Terminal, one of Asia's busiest passenger piers, requires regular inspections to detect any signs of rust, cracks, or structural collapse. The current methods rely on human divers for underwater inspections. Implementing robotic inspection systems could provide a safer approach while enabling more thorough examinations in confined spaces that are challenging for human divers to access. To enable underwater robots to perform comprehensive and efficient inspection tasks, the development of global path planning algorithms for complex marine environments emerges as a technical challenge, as shown in Fig.~\ref{fig:fig1_2}.

Global path-planning methods in complex environments can be categorized into three main types. Classical search-based methods, such as A*~\cite{hart1968formal}, provide resolution optimality but are often limited by high computational and memory demands in high-dimensional spaces. Artificial potential field (APF) methods~\cite{khatib1986real} utilize gradient descent to guide robot motion; however, they are prone to being trapped in local minima. Sampling-based methods, including rapidly exploring random tree (RRT)~\cite{lavalle1998rapidly}, iteratively explore and exploit environmental information to identify globally optimal solutions. An advanced variant, RRT*~\cite{karaman2011sampling}, achieves asymptotic optimality by incrementally reconstructing search trees in continuous space. Its ability to balance exploration and exploitation in high-dimensional complex spaces makes it a widely adopted approach for global path planning in many robotics tasks nowadays. A classic work that improves planning speed by utilizing geometric heuristic regions is Informed RRT*~\cite{gammell2018informed}, which enhances the sampling efficiency by focusing on an \( L_2 \)-informed set, reducing the exploration of irrelevant areas in the state space. Cyl-iRRT*~\cite{yu2023cyl} achieves efficient and safe path planning for autonomous underwater vehicles in 3D environments by focusing the search space within a shrinking cylinder. A recent work~\cite{chen2024rbi} further improves planning efficiency and speed by combining Informed RRT* with bidirectional sampling, enhancing the overall sampling process. Although these heuristic RRT* methods are effective, their performance is limited by the quality of the initial solution, and designing the geometric regions themselves can be quite cumbersome.

In recent years, researchers have started applying neural network methods to improve the performance of RRT*-based algorithms, by learning from a large number of successful optimal path planning cases to generate targeted heuristic regions that enhance sampling efficiency. Among them, two pioneering works stand out. Neural RRT*~\cite{wang2020neural, wang2021deep} uses a network architecture similar to U-Net~\cite{ronneberger2015u}, taking the original map as input and directly predicting heuristic regions for biased sampling by RRT*-based tree methods, greatly enhancing efficiency. MPNet~\cite{qureshi2020motion}, on the other hand, utilizes convolutional neural networks to directly output the optimal path. In recent work, Neural Informed RRT*~\cite{huang2024neural} combines the advantages of traditional informed sets and neural network predictions, using a point cloud representation to further enhance the planning performance of RRT* methods. \textcolor{blue}{Although the aforementioned methods have achieved good results, they are usually applied in map environments that are regular in nature. In addition, the testing environments in these studies are often very similar to the training environments, typically consisting of indoor or laboratory environments with well-defined geometries and known workspace configurations. Research on neural network-assisted RRT* methods for underwater robotics in highly cluttered ocean environments—such as involving underwater pier inspection with irregular structures, aquatic vegetation, and other natural obstacles—is an area that remains missing in the current literature.} This motivates our research.

In this paper, our framework integrates bidirectional search with neural heuristic regions, resulting in a novel neural network tailored for cluttered marine environments that produces high-quality heuristic regions. This integration allows for more efficient path planning by reducing the search space and accelerating convergence. The contributions are multifold:
\begin{enumerate}[label=\arabic*)]
  \item We propose PierGuard, a novel and fast global path planning framework designed for underwater inspection of different pier structures in the ocean. 
  \item We design a novel neural network architecture to generate high-quality heuristic regions, enhancing planning performance in cluttered underwater environments.
  \item We validate the effectiveness and efficiency of our approach through both simulation environments and real-world ocean experiments. The source code will be made publicly available. 
\end{enumerate}

The remainder of this paper is organized as follows. Section II provides a review of related work to establish the context of this study. Section III introduces the key preliminaries and foundational concepts. Section IV elaborates on the proposed methodology in detail. Section V presents the experimental results and analysis. Section VI concludes the paper and outlines directions for future work.

\section{RELATED WORKS}

\subsection{Classical Sampling-based Tree Methods}

The classical Probabilistic Roadmap (PRM)~\cite{kavraki1996probabilistic} is a multi-query sampling-based method that constructs a graph of randomly sampled points in the free space, enabling efficient path finding in complex environments. RRT generates a tree through random sampling for fast exploration, while RRT* enhances this by progressively optimizing the path to achieve asymptotic optimality. RRT\#~\cite{arslan2013use} improves upon RRT* by not only ensuring asymptotic optimality but also providing immediate access to the lowest-cost path information within its spanning tree. The RRT-connect~\cite{kuffner2000rrt} and B-RRT*~\cite{JordanPerez2013} algorithms extend the RRT framework by growing two trees simultaneously from the start and goal points, using a bidirectional search strategy to improve efficiency and convergence. IB-RRT*~\cite{qureshi2015intelligent} algorithm builds upon B-RRT* by incorporating an intelligent sample insertion heuristic, enabling faster convergence to the optimal path. In~\cite{lai2019balancing}, the authors combine incremental multi-query planning with the use of multiple disjointed trees, enhancing efficiency in exploring and optimizing paths. Recent advancements in sampling strategies have introduced various heuristic designs to enhance performance. Informed RRT*~\cite{gammell2018informed} and Guild RRT*~\cite{scalise2023guild} utilize a direct sampling approach confined to the informed set or local sets after identifying a feasible path. Batch Informed Trees (BIT*)~\cite{gammell2020batch}, an anytime and informed sampling-based planner, integrates sampling and heuristic methods to iteratively estimate and explore the problem space. Building upon BIT*, the authors in~\cite{strub2022adaptively} propose Adaptively Informed Trees (AIT*) and Effort Informed Trees (EIT*), two asymptotically optimal approaches that improve planning by dynamically computing and leveraging problem-specific heuristics. More recently, Scalise \textit{et al.}~\cite{scalise2023guild} introduce a local densification technique to optimize the informed set's efficiency by incorporating subsets defined using beacons. Additionally, Reconstructed Bi-directional Informed RRT* (RBI-RRT*)~\cite{chen2024rbi} combines the rapid search capabilities of RRT-connect with the informed sampling strategy of Informed RRT*, offering a robust solution for high-dimensional planning problems. RT-RRT~\cite{cui2024reverse} introduces a reverse tree-guided rapid exploration algorithm that enhances path planning in dynamic environments by combining reverse and forward trees with path optimization. \textcolor{blue}{Although methods based on geometry-inspired heuristic regions have achieved some effectiveness, they suffer from two major drawbacks. First, their performance typically relies on a good initial path solution, meaning a feasible path that is not only collision-free but also reasonably close to the optimal, as its quality directly affects the effectiveness of the constructed heuristic region. Second, they require tuning geometric parameters for different maps, such as the shape and size of the informed region or the placement and number of beacon nodes.}

\subsection{Learning-enhanced Sampling-based Tree Methods}

With the advancement of deep learning technology, many researchers have begun integrating neural networks with the RRT* method to predict heuristic regions, initial paths, or waypoints, thereby improving overall path planning performance. Kuo \textit{et al.}~\cite{kuo2018deep} integrate a sequence model with a sampling-based planner, enabling the planner's subsequent moves and states to be influenced by observations. Chiang \textit{et al.}~\cite{chiang2019rl} introduced the RL-RRT method, which leverages deep deterministic policy gradients in conjunction with upper-layer sampling-based planning techniques to accomplish planning. MPNet~\cite{qureshi2020motion} is a neural motion planner that learns heuristics for efficient, near-optimal path planning in diverse environments. Wang \textit{et al.}~\cite{wang2020neural} present an end-to-end pixel-based path planning framework using Convolutional Neural Networks (CNN) to directly predict promising regions for optimal paths without preprocessing and then extend it to 3D environment~\cite{wang2021deep}. Similarly, Generative Adversarial Networks (GAN) have been applied to enable non-uniform sampling in RRT*-like algorithms~\cite{li2021efficient}. Neural Informed RRT*~\cite{huang2024neural} combines rule-based informed sampling with learning-based methods to enhance the efficiency of motion planning. It leverages point cloud representations of free states to guide the sampling process through a two-step approach. In~\cite{wang2024miner}, the authors further improve the connectivity and safety of the generated heuristic regions, thereby improving the overall performance of path planning. Compared to geometry-inspired heuristic regions, neural network-based heuristic region methods demonstrate significant improvements in aspects such as finding high-quality initial solutions. \textcolor{blue}{However, existing neural network-based methods suffer from two main limitations. First, the generated neural heuristic regions often lack consideration of important geometric properties of the environment, such as obstacle shape and spatial layout, which limits their adaptability in cluttered scenarios. Second, most existing approaches adopt unidirectional search strategies, without leveraging bidirectional sampling to further improve planning efficiency.}

\section{PRELIMINARIES}

\subsection{Optimal Path Planning}

Let $X \subseteq \mathbb{R}^n$ represent the state configuration space, $X_{\mathrm{obs}} \subset X$ denote the space occupied by obstacles, and $X_{\mathrm{free}} = closure\left( X \backslash X_{\mathrm{obs}} \right)$ correspond to the unoccupied space. The initial condition and goal condition are represented as $x_{\mathrm{init}} \in X_{\mathrm{free}}$ and $x_{\mathrm{goal}} \in X_{\mathrm{goal}} = \left\{ x \in X \mid \left\| x - x_{\mathrm{goal}} \right\| < radius \right\}$, respectively~\cite{lavalle2006planning}. 

\textit{Feasible Path Planning Problem}: Consider the triplet $\left( X_{\mathrm{free}}, x_{\mathrm{init}}, X_{\mathrm{goal}} \right)$ and the continuous time $t$, find a feasible path comprised of collision-free states $\varsigma \left( t \right) : \left[ 0,T \right] \rightarrow X_{\mathrm{free}}$ such that $\varsigma \left( 0 \right) = x_{\mathrm{init}}$, and $\varsigma \left( T \right) \in X_{\mathrm{goal}}$.

\textit{Optimal Path Planning Problem}: Let $\varSigma$ denote the set of all feasible paths, and let $c\left( \varsigma \right)$ denote the cost function for path evaluation. 
\begin{equation}
\begin{aligned}
    & c\left( \varsigma ^* \right) = \underset{\varsigma \in \varSigma}{\arg\min} \, c\left( \varsigma \right) \\
    & s.t. \varsigma \left( 0 \right) = x_{\mathrm{init}}, \varsigma \left( T \right) \in X_{\mathrm{goal}}, \varsigma \left( t \right) \in X_{\mathrm{free}}, \forall t \in \left[ 0,T \right],
\end{aligned}
\end{equation}
where the cost function between two states is defined as:
\begin{equation}
c(x_1,x_2)=\beta _1\parallel x_2-x_1\parallel _2+\beta _2\mathrm{arc}\cos \left( \frac{\overrightarrow{v_1}\cdot \overrightarrow{x_1x_2}}{\left| \overrightarrow{v_1} \right|\left| \overrightarrow{x_1x_2} \right|} \right),
\end{equation}
and $\overrightarrow{v_1}$ is the linear velocity vector, $\overrightarrow{x_1x_2}$ is the vector from $x_1$ to $x_2$.

\subsection{Rapidly-exploring Random Trees}
The RRT*~\cite{karaman2011sampling} algorithm incrementally builds a tree structure in the search space to find a near-optimal path to a target. The algorithm initializes with a tree $T = (V, E)$, where $V$ is the set of vertices and $E$ is the set of edges, starting at the initial state $x_{\text{init}}$. At each iteration, a random sample $x_{\text{rand}}$ is drawn from the state space, and the nearest tree vertex $x_{\text{nearest}}$ is identified based on a distance metric $c(x_1, x_2)$. A new state $x_{\text{new}}$ is generated by steering from $x_{\text{nearest}}$ towards $x_{\text{rand}}$ using the steering function.

The feasibility of the new connection $(x_{\text{nearest}}, x_{\text{new}})$ is checked, ensuring that no obstacles obstruct the path. If feasible, $x_{\text{new}}$ is added to the tree by extending $T$ and updating $E$. The rewiring step optimizes the tree locally by minimizing the cost of connections around $x_{\text{new}}$. If $x_{\text{new}}$ reaches the goal region $X_{\mathrm{goal}}$, the algorithm returns the tree $T$ containing a path to the goal. RRT* ensures asymptotic optimality by iteratively refining the solution path and minimizing the cumulative cost function over the tree structure. 

\section{METHODOLOGY}

\begin{figure*}[ht!]
    \centering
    \includegraphics[width=\textwidth]{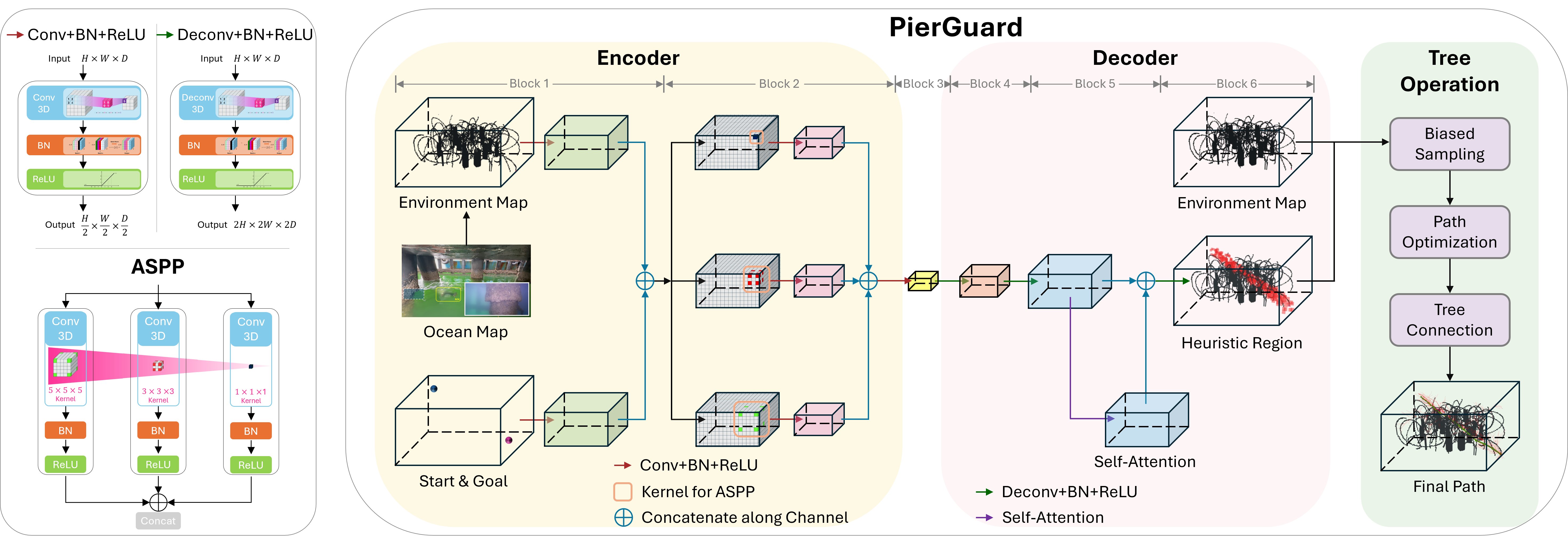}
    \caption{\textcolor{blue}{An overview of the proposed PierGuard framework.}}
    \label{fig:frame}
\end{figure*}

An overview of our proposed PierGuard framework is shown in Fig.~\ref{fig:frame}. The left half showcases our high-quality heuristic region generation, specifically designed for cluttered underwater environments. The right half integrates bidirectional sampling with neural heuristic regions in a sampling-based path-finding method.

\subsection{Design for cluttered Environment}\label{sec:network}

\subsubsection{Neural Network Design}

We propose a neural network architecture for feature extraction in cluttered underwater environments. The input consists of two three-dimensional grid maps, denoted as \(\mathcal{E}_1\) (start/goal positions) and \(\mathcal{E}_2\) (obstacles and free space), which are separately processed using convolutional encoders to extract spatial features before being concatenated.  

As shown in the encoder section of Fig.~\ref{fig:frame}, to enhance the network’s ability to capture multi-scale spatial dependencies, an Atrous Spatial Pyramid Pooling (ASPP)~\cite{chen2018encoder} module is applied, allowing for effective feature aggregation across different receptive fields. The encoded feature representation of the environment \(\mathcal{E}\) is progressively downsampled, forming a compact latent representation while preserving essential structural information.  

The decoder reconstructs the heuristic region \(\mathcal{H}\) using a series of deconvolution layers, each followed by batch normalization (BN) and ReLU activation to ensure stable gradient flow. Self-attention mechanisms~\cite{zhang2019self} are incorporated at intermediate decoding stages to refine long-range dependencies and adapt to the complex spatial structures of underwater terrains, as shown in the decoder section of Fig.~\ref{fig:frame}.

The final output \(\mathcal{H}\) retains the same resolution as \(\mathcal{E}\), providing an informative heuristic representation to facilitate efficient motion planning in challenging cluttered environments.

\subsubsection{Loss Function Design}

The quality of the heuristic region directly affects the path-planning algorithm. Compared to the traditional Binary Cross-Entropy (BCE) loss~\cite{wang2021deep}, our proposed formulation increases the weight of pixels near the ground-truth path to enhance learning in critical areas:

\begin{equation}
    \mathcal{L}_{\text{path}} = - \sum_i w_i \left[ p_{\text{gt},i} \log(p_{o,i}) + (1 - p_{\text{gt},i}) \log(1 - p_{o,i}) \right],
\end{equation}
where \( p_{\text{gt},i} \) represents the ground-truth probability of pixel \( i \) belonging to the ground-truth path, while \( p_{o,i} \) is the predicted probability. \textcolor{blue}{The weighting factor \( w_i = 1 + \lambda d_i^{-1} \) is determined by the minimum distance \( d_i \) from the current pixel to the ground-truth path, and \( \lambda \) controls the weight (set to 10 based on empirical studies). This design prefers exploitation by focusing on critical regions, while still allowing moderate exploration around the path.} This formulation directs the model's attention toward regions near the ground-truth path while reducing the influence of distant background pixels.

The heuristic region should also maintain a reasonable geometric shape with strong connectivity and the predicted region should be topologically consistent with the ground-truth path generated by the A* algorithm. The Hausdorff distance can be used to measure the shape similarity between the two:

\begin{equation}
    \mathcal{L}_{\text{haus}} = \max \left\{ \max_{x \in P_o} \min_{y \in P_{\text{gt}}} d(x,y), \max_{y \in P_{\text{gt}}} \min_{x \in P_o} d(x,y) \right\},
\end{equation}
where \( P_o \) is the set of predicted heuristic region points, \( P_{\text{gt}} \) represents the set of ground-truth path points, and \( d(x, y) \) denotes the Euclidean distance between two points. Unlike~\cite{wang2024miner}, which focuses solely on improving connectivity, our formulation not only reduces excessive isolated points to enhance connectivity but also improves shape alignment with the ground-truth path.

The total loss function is defined as:
\begin{equation}
    \mathcal{L}_{\text{total}} = \alpha_1\mathcal{L}_{\text{path}} + \alpha_2\mathcal{L}_{\text{haus}},
\end{equation}
where $\alpha_1$ and $\alpha_2$ are coefficients. \textcolor{blue}{Together, these two terms improve both local accuracy and global structure, which are essential for downstream planning.}

\subsection{PierGuard Algorithm}
The overall structure of the proposed PierGuard algorithm is presented in Alg.~\ref{alg:ocean_rrt}. PierGuard integrates three key components: the generation of neural heuristic regions combined with biased sampling to guide the search process [Alg.~\ref{alg:biased_sampling}], reconnection techniques for path optimization [Alg.~\ref{alg:path_opt}], and tree connection and optimization strategies to enhance overall planning performance [Alg.~\ref{alg:tree_connection}].

\begin{algorithm}
    \renewcommand{\algorithmicrequire}{\textbf{Input:}}
    \renewcommand{\algorithmicensure}{\textbf{Output:}}
    \caption{PierGuard Algorithm}
    \label{alg:ocean_rrt}
    \begin{algorithmic}[1]
        \REQUIRE Map $\mathcal{E}$, $x_{\mathrm{init}}, x_{\mathrm{goal}}$
        \ENSURE Tree $G = \left( V_{N}^{\text{PierGuard}},E \right)$
        \STATE Vertex $V_{N}^{\text{PierGuard}} \gets \{x_{\text{init}}, x_{\text{goal}}\}$, Edge $E \gets \emptyset$
        \STATE Forward $T_a \gets \{x_{\text{init}}, E\}$, Backward $T_b \gets \{x_{\text{goal}}, E\}$
        \STATE $c_{\text{best}} \gets \infty$, $\varsigma_{\text{best}} \gets \emptyset$
        \STATE Heuristics $\mathcal{H}\gets \textbf{Network}\left( x_{\mathrm{init}}, x_{\mathrm{goal}}, \mathcal{E} \right)$
        \FOR{$i \gets 1$ to $N$}
            \STATE \textbf{NeuralSample} ($\mathcal{E}, \mathcal{H}$)
            \STATE \textbf{PathOpt} ($x_{\mathrm{new}}, X_{\mathrm{near}}$)
            \STATE \textbf{TreeConnect} $(L_{\text{near}}, T_a, T_b)$
        \ENDFOR
        \RETURN $G$
    \end{algorithmic}
\end{algorithm}

\subsubsection{Biased Sampling and Extension}
First, the algorithm initializes the start point $x_{\mathrm{init}}$, goal point $x_{\mathrm{goal}}$, map information $\mathcal{E}$, and the tree structures $G, T_a, T_b$ [Alg.~\ref{alg:ocean_rrt} Lines 1-3]. Here, $\varsigma_{\text{best}}$ represents the current optimal path, and $c_{\text{best}}$ denotes the corresponding cost value associated with this optimal path. Subsequently, a neural network specifically designed for cluttered environments is employed to generate high-quality heuristic regions $\mathcal{H}$ [Alg.~\ref{alg:ocean_rrt} Lines 4], which will be detailed in Sec.~\ref{sec:network}. When sampling new nodes, our method primarily samples within the heuristic regions while ensuring a certain probability of uniformly sampling in the global space to maintain probabilistic completeness [Alg.~\ref{alg:biased_sampling} Lines 1-5], where $\mu$ is a pre-set threshold that governs the sampling distribution. Finally, the algorithm returns a triplet ($x_{\mathrm{nearest}}, x_{\mathrm{new}}, X_{\mathrm{near}}$).

\begin{algorithm}
    \renewcommand{\algorithmicrequire}{\textbf{Function:}}
    \caption{Biased Sampling and Extension}
    \label{alg:biased_sampling}
    \begin{algorithmic}[1]
        \REQUIRE \textbf{NeuralSample} ($\mathcal{E}, \mathcal{H}$)
        \IF{\textbf{Rand()} $< \mu$}
            \STATE $x_{\mathrm{rand}} \gets \textbf{Uniform}(\mathcal{E})$
        \ELSE
            \STATE $x_{\mathrm{rand}} \gets \textbf{NonUniform}(\mathcal{H})$
        \ENDIF
        \STATE $x_{\mathrm{nearest}} \gets \textbf{Nearest}(T_a, x_{\mathrm{rand}})$
        \STATE $x_{\mathrm{new}} \gets \textbf{Steer}(x_{\mathrm{nearest}}, x_{\mathrm{rand}})$
        \STATE $X_{\mathrm{near}} \gets \textbf{Near}(T_a, x_{\mathrm{new}})$
        \RETURN ($x_{\mathrm{nearest}}, x_{\mathrm{new}}, X_{\mathrm{near}}$)
    \end{algorithmic}
\end{algorithm}

\subsubsection{Path Optimization and Re-connect}
Initially, the triplet $(c_{\text{near}}, x_{\text{near}}, \varsigma_{\text{near}})$ is computed and added by traversing the $x_{\text{near}}$ set [Alg.~\ref{alg:path_opt} Line 1-7]. Subsequently, the algorithm iterates through the candidate paths to update the vertex and edge sets of the tree [Alg.~\ref{alg:path_opt} Line 8-17]. Finally, the algorithm performs another traversal of the candidate paths, attempting to reconnect branches if a new path is found to be superior to the original one [Alg.~\ref{alg:path_opt} Line 18-26].

\begin{algorithm}
    \renewcommand{\algorithmicrequire}{\textbf{Function:}}
    \caption{Path Optimization and Re-connect}
    \label{alg:path_opt}
    \begin{algorithmic}[1]
        \REQUIRE \textbf{PathOpt} ($x_{\mathrm{new}}, X_{\mathrm{near}}$)
            \STATE $L_{\text{near}} \gets \emptyset$
            \FORALL{$x_{\text{near}} \in X_{\text{near}}$}
                \STATE $\varsigma_{\text{near}} \gets \textbf{Steer}(x_{\text{near}}, x_{\text{new}})$
                \STATE $c_{\text{near}} \gets \textbf{Cost}(x_{\text{near}}) + \textbf{Cost}(O_{\text{near}})$
                \STATE $L_{\text{near}} \gets L_{\text{near}} \cup \{(c_{\text{near}}, x_{\text{near}}, \varsigma_{\text{near}})\}$
            \ENDFOR
            \STATE $L_{\text{near}}.\textbf{Sort}()$
            \FORALL{$(c_{\text{near}}, x_{\text{near}}, \varsigma_{\text{near}}) \in L_{\text{near}}$}
                \IF{$\textbf{CollisionFree}(\varsigma_{\text{near}})$}
                    \IF{$c_{\text{near}} + \textbf{CostToGo}(x_{\text{new}}) < c_{\text{best}}$}
                        \STATE $x_{\text{min}} \gets x_{\text{near}}$
                        \STATE $V \gets V \cup \{x_{\text{new}}\}$
                        \STATE $E \gets E \cup \{(x_{\text{min}}, x_{\text{new}})\}$
                        \STATE \textbf{break}
                    \ENDIF
                \ENDIF
            \ENDFOR
            \FORALL{$(c_{\text{near}}, x_{\text{near}}, \varsigma_{\text{near}}) \in L_{\text{near}}$}
                \IF{$\textbf{Cost}(x_{\text{new}}) + c_{\text{near}} < \textbf{Cost}(x_{\text{near}})$}
                    \IF{$\textbf{CollisionFree}(\varsigma_{\text{near}})$}
                        \STATE $x_{\text{oldparent}} \gets \textbf{Parent}(E, x_{\text{near}})$
                        \STATE $E \gets E \setminus \{(x_{\text{oldparent}}, x_{\text{near}})\}$
                        \STATE $E \gets E \cup \{(x_{\text{new}}, x_{\text{near}})\}$
                    \ENDIF
                \ENDIF
            \ENDFOR
    \end{algorithmic}
\end{algorithm}

\subsubsection{Tree Connection and Optimization}
First, the algorithm attempts to connect the forward and backward trees. If the newly formed path resulting from this connection is superior, a global update is performed [Alg.~\ref{alg:tree_connection} Line 1-6]. Subsequently, the algorithm performs operations on the tree such as vertex contraction, removal of high-cost paths, and swapping [Alg.~\ref{alg:tree_connection} Line 7-8], similar to those described in~\cite{JordanPerez2013}. Finally, the algorithm returns the tree structure if the loop count is reached or if the optimal path is found.

\begin{algorithm}
    \renewcommand{\algorithmicrequire}{\textbf{Function:}}
    \caption{Tree Connection and Optimization}
    \label{alg:tree_connection}
    \begin{algorithmic}[1]
        \REQUIRE \textbf{TreeConnect} $(L_{\text{near}}, T_a, T_b)$
        \STATE $x_{\mathrm{connect}} \gets \textbf{Nearest}(T_b, x_{\mathrm{new}})$
        \STATE $(c_{\mathrm{sol}}, \varsigma_{\mathrm{sol}}) \gets \textbf{ConnectGraphs}(T_b, x_{\mathrm{connect}}, x_{\mathrm{new}})$
        \IF{$c_{\mathrm{sol}} < c_{\mathrm{best}}$}
            \STATE $c_{\mathrm{best}} \gets c_{\mathrm{sol}}$
            \STATE $\varsigma_{\mathrm{best}} \gets \varsigma_{\mathrm{sol}}$
        \ENDIF
        \STATE $\textbf{BranchAndBound}(T_a, T_b)$
        \STATE $\textbf{SwapTrees}(T_a, T_b)$
        \RETURN $G$
    \end{algorithmic}
\end{algorithm}

\subsection{Algorithm Analysis}

In this section, we demonstrate that PierGuard inherits both the probabilistic completeness and asymptotic optimality properties from the original RRT*-based algorithm. In addition, we analyze the complexity of the PierGuard algorithm.

\textit{Lemma 1 (Probabilistic Completeness of PierGuard~\cite{karaman2011sampling}):} The PierGuard algorithm is probabilistically complete, i.e., for any robustly feasible planning problem, the following holds:
\begin{equation}
    \lim_{N\rightarrow \infty} \mathbb{P}\left( V_{N}^{\text{PierGuard}}\cap {X} _{\mathrm{goal}}\ne \emptyset \right) =1.
\end{equation}
where $V_{N}^{\text{PierGuard}}$ is the set of vertices after $N$ iterations for PierGuard algorithm and ${X} _{\mathrm{goal}}$ is the goal region.

\begin{proof}
We first define $V_{N}^{\text{RRT}^*}$ as the final set of vertices for RRT* when the number of sampling nodes is $N$. According to~\cite{karaman2011sampling}, the RRT* algorithm enjoy probabilistic completeness. When $N\rightarrow \infty$, it follows that $V_{N}^{\text{PierGuard}}=V_{N}^{\text{RRT}^*}$. Since PierGuard ensures a connected graph, it possesses the same probabilistic completeness property as RRT*.
\end{proof}

\textit{Lemma 2 (Asymptotic Optimality of PierGuard~\cite{karaman2011sampling}):} The PierGuard algorithm is asymptotically optimal, i.e., for any path planning triplet $\left( {X} _{\mathrm{free}}, x_{\mathrm{init}}, {X} _{\mathrm{goal}} \right)$, optimal cost $c\left( \zeta ^* \right)$ and the minimum cost set $C_{\text{N}}$ obtained by tree query through our algorithm, the following holds:
\begin{equation}
    \mathbb{P}\left( \left\{ \lim_{N\rightarrow \infty} \sup C_{N} = c\left( \zeta^* \right) \right\} \right) = 1,
\end{equation}
if and only if we let the searching radius $\eta$ satisfy the following condition:
\begin{equation}
    \eta >\left( 2\left( 1+\frac{1}{m} \right) \right) ^{\frac{1}{m}}\left( \frac{\mathcal{M} \left( {X} _{\mathrm{free}} \right)}{\xi _d} \right) ^{\frac{1}{m}},
\end{equation}
where $m$ is the dimension of state space, $\mathcal{M} \left( \mathcal{X} _{\mathrm{free}} \right)$ is the Lebesgue measure of $\mathcal{X} _{\mathrm{free}}$ and $\xi _d$ is the volume of the unit ball which can all be calculated based on the environment in advance.

\begin{proof}
PierGuard is an improvement to the sampling strategy of RRT* and B-RRT*. It does not alter the extension and rewire processes of them. Then the proof follows from Theorem 38 in~\cite{karaman2011sampling} and Theorem 2 in~\cite{JordanPerez2013}, provided that the search radius satisfies the above conditions.
\end{proof}

\textit{Lemma 3 (Computational Complexity Ratio of PierGuard and RRT-connect~\cite{JordanPerez2013}):} There exists a constant $\delta$ such that the following holds:
\begin{equation}
\lim_{N\rightarrow \infty} \sup \mathbb{E}\left[\frac{M^{\text{PierGuard}}_N}{M^{\text{RRT-connect}}_N}\right] \leq \delta,
\end{equation}
where $M^{\text{PierGuard}}_N$ and $M^{\text{RRT-connect}}_N$ are the total computational steps for PierGuard and RRT-connect algorithms.

\begin{proof}
The proof follows from Theorem 18 in~\cite{karaman2011sampling} and Theorem 3 in~\cite{JordanPerez2013} with slight modifications based on Alg.~\ref{alg:biased_sampling}. In fact, PierGuard and RRT-connect include an additional step in each iteration that attempts to connect the two trees. Assuming the extension step size parameter is sufficiently large, the computational cost of this additional step is equivalent to that of a single-tree extension iteration. This comparison indicates that although PierGuard has enhanced asymptotic optimality features, its computational overhead does not significantly increase compared to RRT-connect.
\end{proof}

\section{EXPERIMENTAL RESULTS AND ANALYSIS}

\subsection{Experiment Setup}
We construct a dataset with approximately 29,874 samples from successful path-finding scenarios in a 3D cluttered ocean environment, which is designed to mimic real-world conditions. \textcolor{blue}{In our framework, collision avoidance is primarily ensured by inheriting the obstacle-checking mechanism of the RRT* algorithm. Specifically, we perform collision checking by interpolating along the line segment between a newly sampled node and its nearest neighbor in the tree, and rejecting any connection that intersects with occupied cells in the environment map.} About 80\% of the samples are used for training and validation, while the remaining 20\% form a test set generated on unseen maps to evaluate generalization. \textcolor{blue}{Ground truth paths are obtained using the A* algorithm and post-processed using dilation to improve neural network recognition. A* provides optimal and efficient solutions on grid maps and has been used in prior learning-based planners~\cite{wang2020neural}. The goal is to obtain high-quality paths for supervision, and therefore, the specific choice of planner is flexible.} \textcolor{blue}{We use Python 3.8.10, PyTorch 1.13.1, and an NVIDIA RTX 3070 Ti for model training and inference. The model is trained for 50 epochs and the training time is 1.2 hours. For a map of size 64x64x64, the inference time is about 30ms, which is enough for real-time path planning. In our experiments, we set \( \alpha_1 = 1.0 \) and \( \alpha_2 = 0.1 \), based on ablation studies on a validation set to balance local pixel-level accuracy and global shape similarity.}

\subsection{Simulation Experiment}

\subsubsection{Geometric Simulation}

\begin{figure}[ht]
    \centering
    \begin{subfigure}{0.49\columnwidth}
        \includegraphics[width=\textwidth]{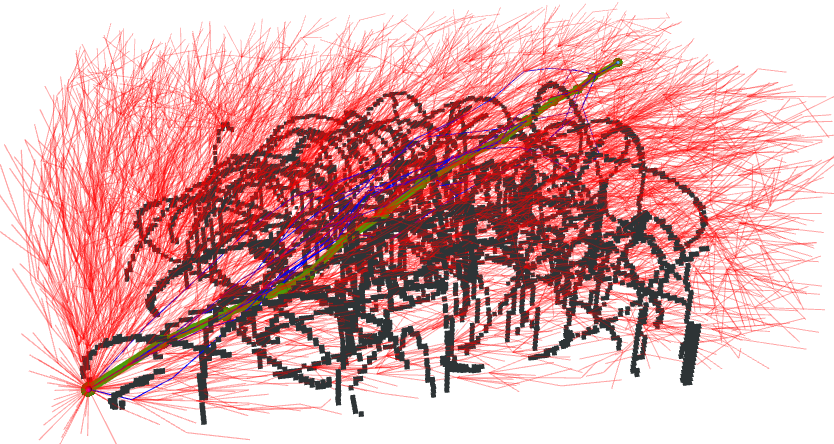}
        \caption{RRT*.}
        \label{fig:simu1_rrt}
    \end{subfigure}
    \begin{subfigure}{0.49\columnwidth}
        \includegraphics[width=\textwidth]{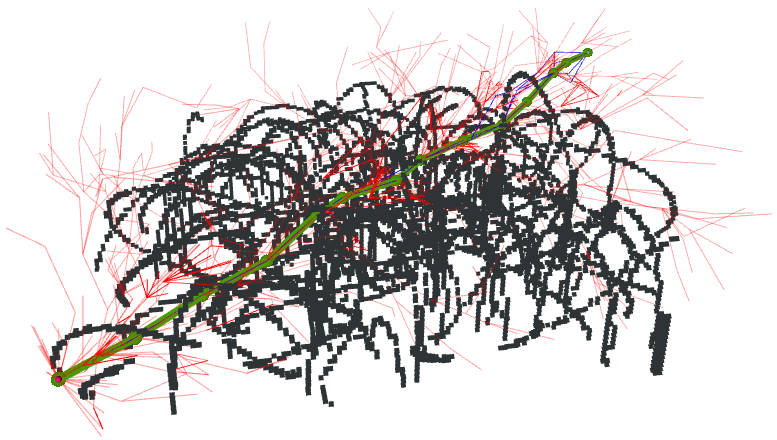}
        \caption{Our method.}
        \label{fig:simu1_ours}
    \end{subfigure}
    \caption{Visualization of RRT* and our method on map 1.}
    \label{fig:simu1}
\end{figure}

\begin{figure}[ht]
    \centering
    \begin{subfigure}{0.49\columnwidth}
        \includegraphics[width=\textwidth]{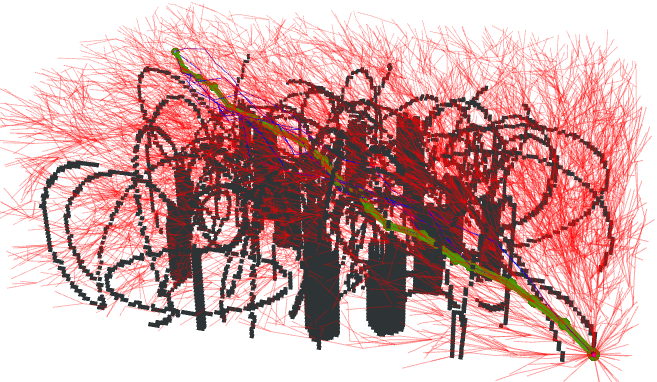}
        \caption{RRT*.}
        \label{fig:simu2_rrt}
    \end{subfigure}
    \begin{subfigure}{0.49\columnwidth}
        \includegraphics[width=\textwidth]{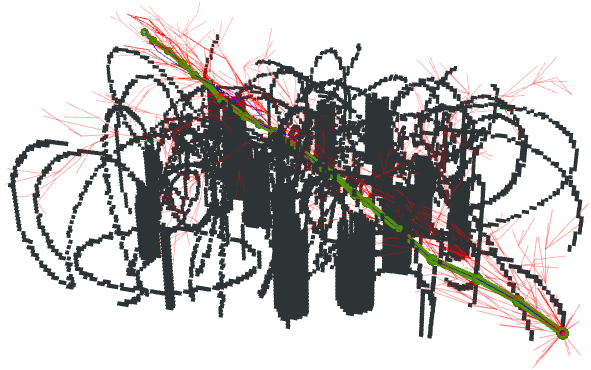}
        \caption{Our method.}
        \label{fig:simu2_ours}
    \end{subfigure}
    \caption{Visualization of RRT* and our method on map 2.}
    \label{fig:simu2}
\end{figure}

On the Ubuntu 18.04 platform, we develop a simulation platform utilizing the Robot Operating System (ROS), where maps are pre-built and stored in the form of occupancy grids to enable obstacle collision detection. In the paper, we select two representative maps and visualize the path planning results, as shown in Fig.~\ref{fig:simu1} - Fig.~\ref{fig:simu2}. In the visualization, the black grid represents obstacle areas, while the white grid indicates free space. The light red edges represent the final growth of the tree, and the green nodes and edges correspond to the final path points and path segments.

In the first experimental scenario, the obstacles primarily consist of circular obstacles with radii ranging between 0.4 and 2.5 and heights between 0.5 and 7.5. As shown in Fig.~\ref{fig:simu1}, the traditional RRT* method generates a significant number of invalid samples in the global space, resulting in substantial time consumption to find both the initial solution and the optimal solution. In contrast, our method leverages bidirectional search within a high-quality heuristic region, significantly reducing invalid sampling and tree operations, thereby enabling rapid discovery of both the initial solution and the optimal solution.

In the second experimental scenario, in addition to circular obstacles, there are cylindrical obstacles with heights ranging from 5.5 to 10.5, closely resembling real-world pier environments, as shown in Fig~\ref{fig:simu2}. \textcolor{blue}{Similarly, our method qualitatively illustrates the advantage by highlighting the significantly reduced number of tree nodes required for planning compared to traditional methods, as visually reflected by the much smaller red regions.}

\subsubsection{Physical Model-based Simulation}

\begin{figure}[ht]
    \centering
    \begin{subfigure}{0.49\columnwidth}
        \includegraphics[width=\textwidth]{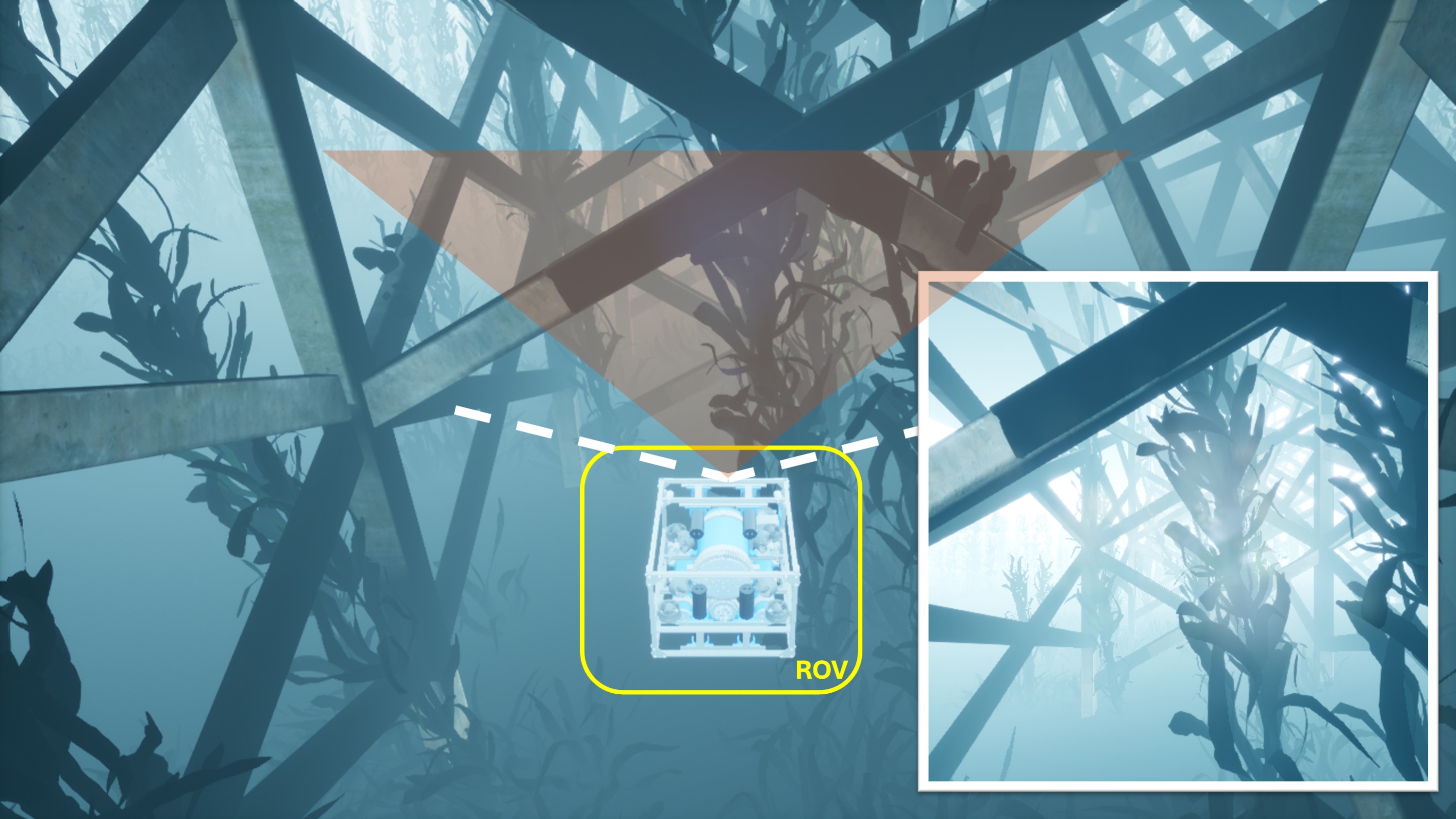}
        \caption{Physical simulation 1.}
        \label{fig:physimu1}
    \end{subfigure}
    \begin{subfigure}{0.49\columnwidth}
        \includegraphics[width=\textwidth]{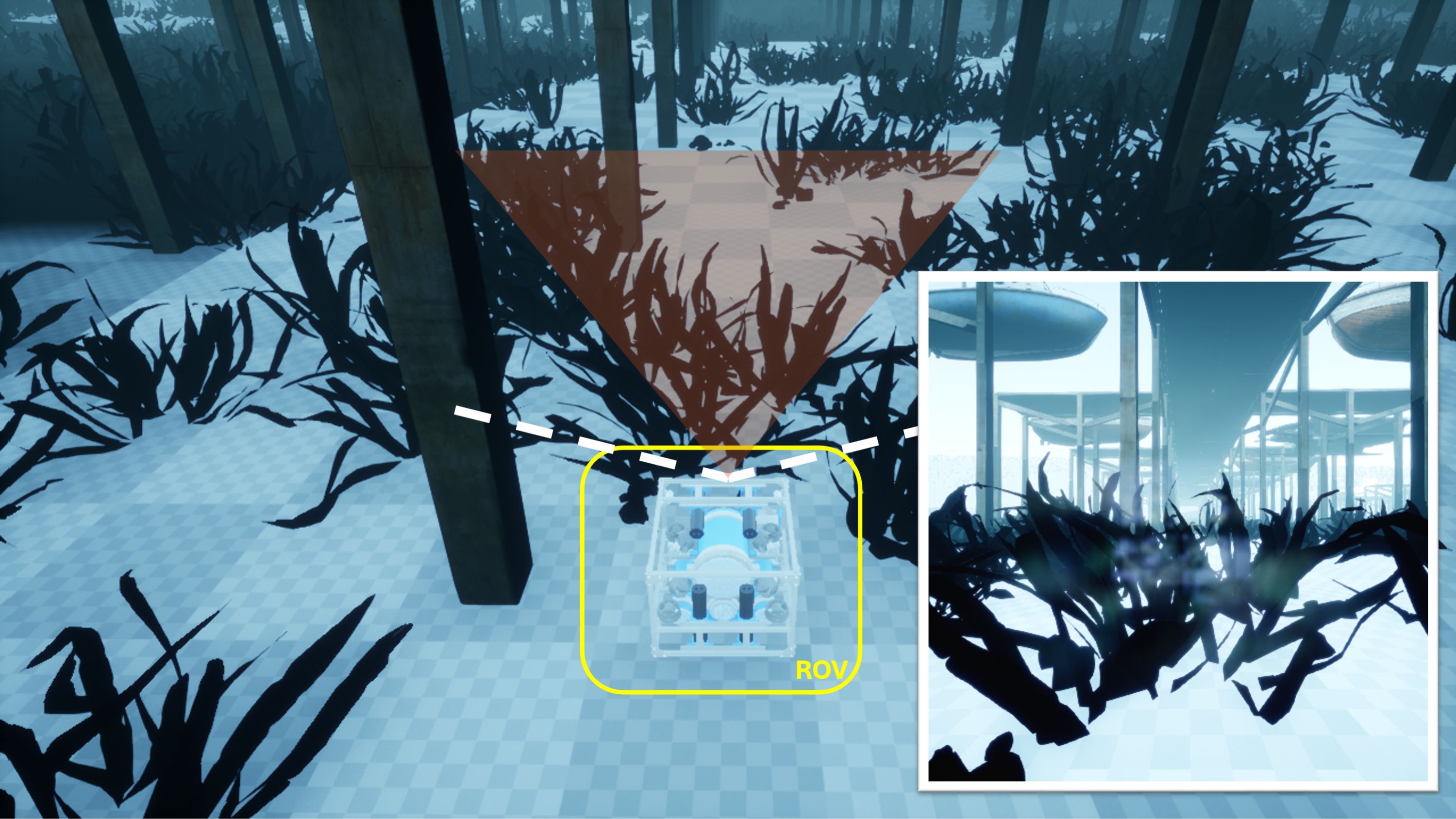}
        \caption{Physical simulation 2.}
        \label{fig:physimu2}
    \end{subfigure}
    \caption{\textcolor{blue}{Physical model-based simulation experiments.}}
    \label{fig:physimu}
\end{figure}

\textcolor{blue}{In order to better simulate real complex scenes, we conduct inspection experiments on pier environments in a simulator with a physical engine~\cite{potokar2024holoocean}.} \textcolor{blue}{As shown in Fig.~\ref{fig:physimu}, we select two representative experimental scenarios that feature highly cluttered pier bottom environments, along with natural seagrass as obstacles. Our path planner quickly generates an optimal path in such environments. While executing the path, the robot performs inspections of the pier bottom, and real-time detection images appear in the thumbnails as illustrated.}

\subsection{Real-world Marine Experiment}

\subsubsection{Field Test}

\begin{figure}[htb]
    \centering
    \begin{subfigure}{0.492\columnwidth}
        \includegraphics[width=\textwidth]{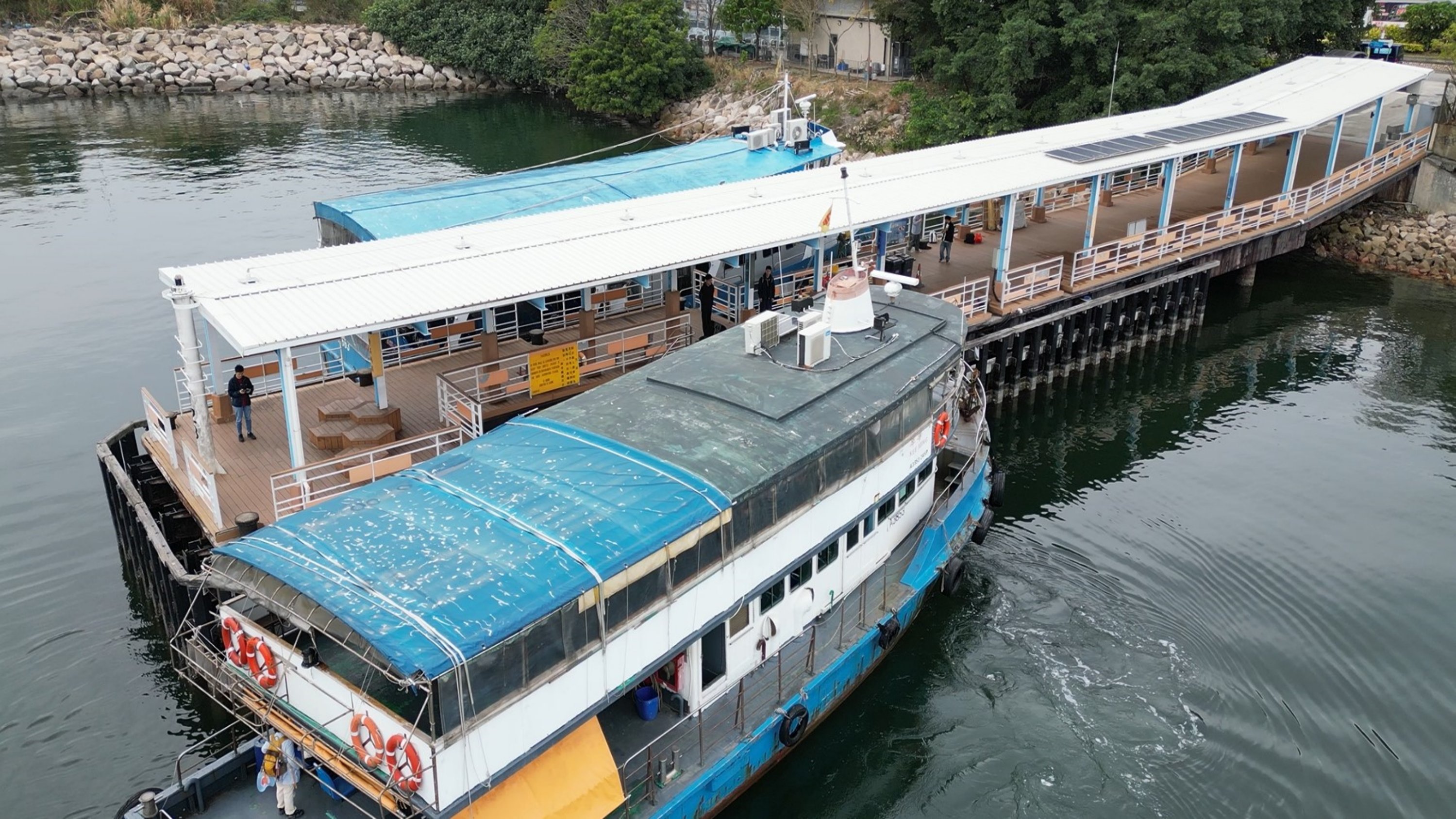}
        \caption{Ma Liu Shui Pier.}
        \label{fig:ma1}
    \end{subfigure}
    \begin{subfigure}{0.492\columnwidth}
        \includegraphics[width=\textwidth]{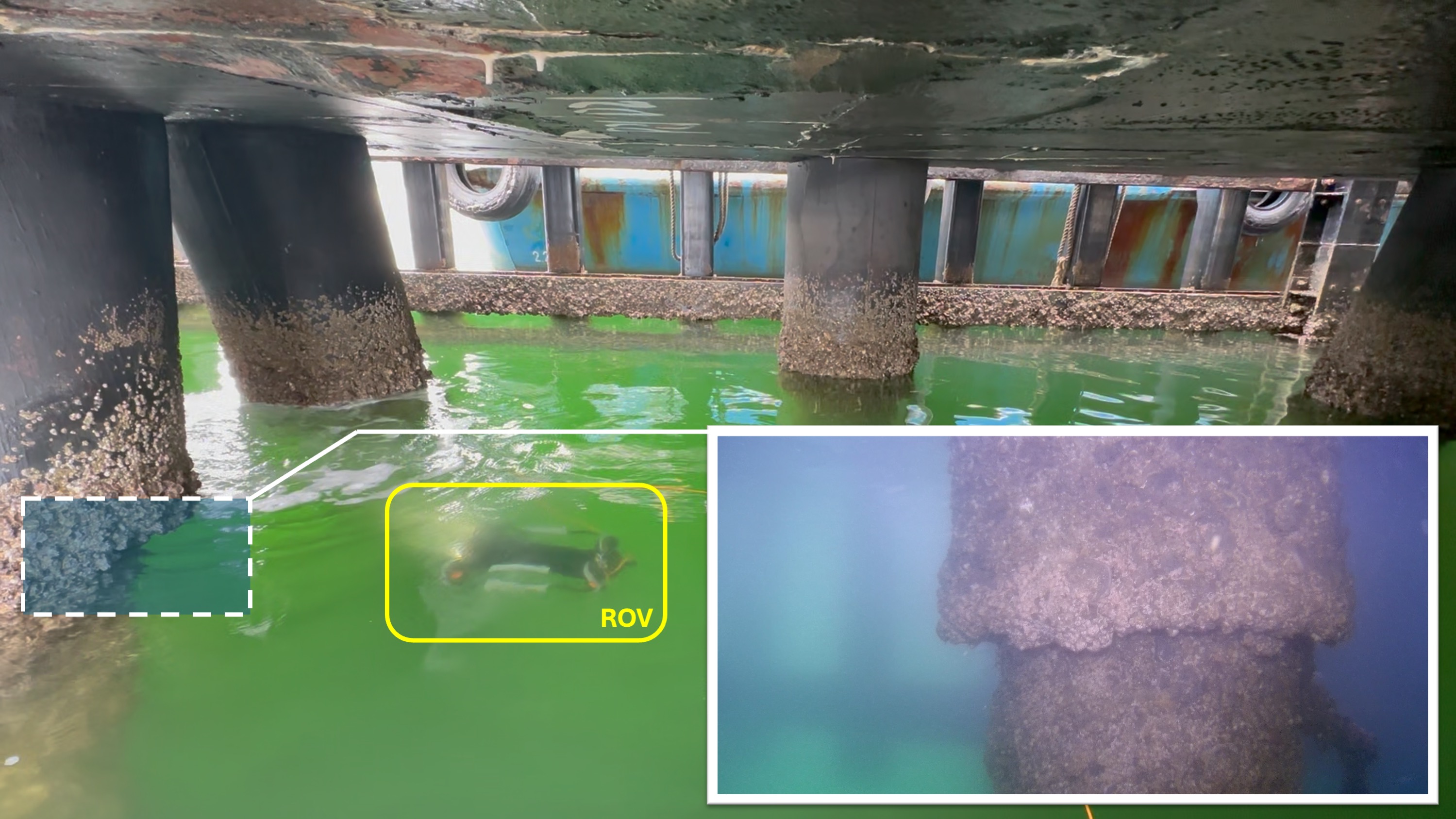}
        \caption{Underwater inspection.}
        \label{fig:ma2}
    \end{subfigure}
    \caption{Real ocean experiment 1 in Hong Kong.}
    \label{fig:maliaoshui}
\end{figure}

We conduct field experiments in real marine conditions in Hong Kong, first selecting the Ma Liu Shui Pier, one of the most renowned piers in the region, for our comprehensive testing, as shown in Fig.~\ref{fig:maliaoshui}. An aerial view of the Ma Liu Shui Pier captured by a drone is presented in Fig.~\ref{fig:ma1}, showcasing the docking area for vessels. The image illustrates that the pier's foundation is supported by pillars arranged in various orientations. Fig.~\ref{fig:ma2} shows an on-site photograph of the pier's submerged structure, where the underwater robot is highlighted with a yellow bounding box. Our path planning methodology demonstrates its capability to efficiently generate a global path in such complex underwater environments, facilitating subsequent inspection tasks. The inset in the lower right corner provides a view of the pier's structural column, showing corrosion and extensive oyster colonization, which substantiates the importance of our inspection mission.

\begin{figure}[htb]
    \centering
    \begin{subfigure}{0.492\columnwidth}
        \includegraphics[width=\textwidth]{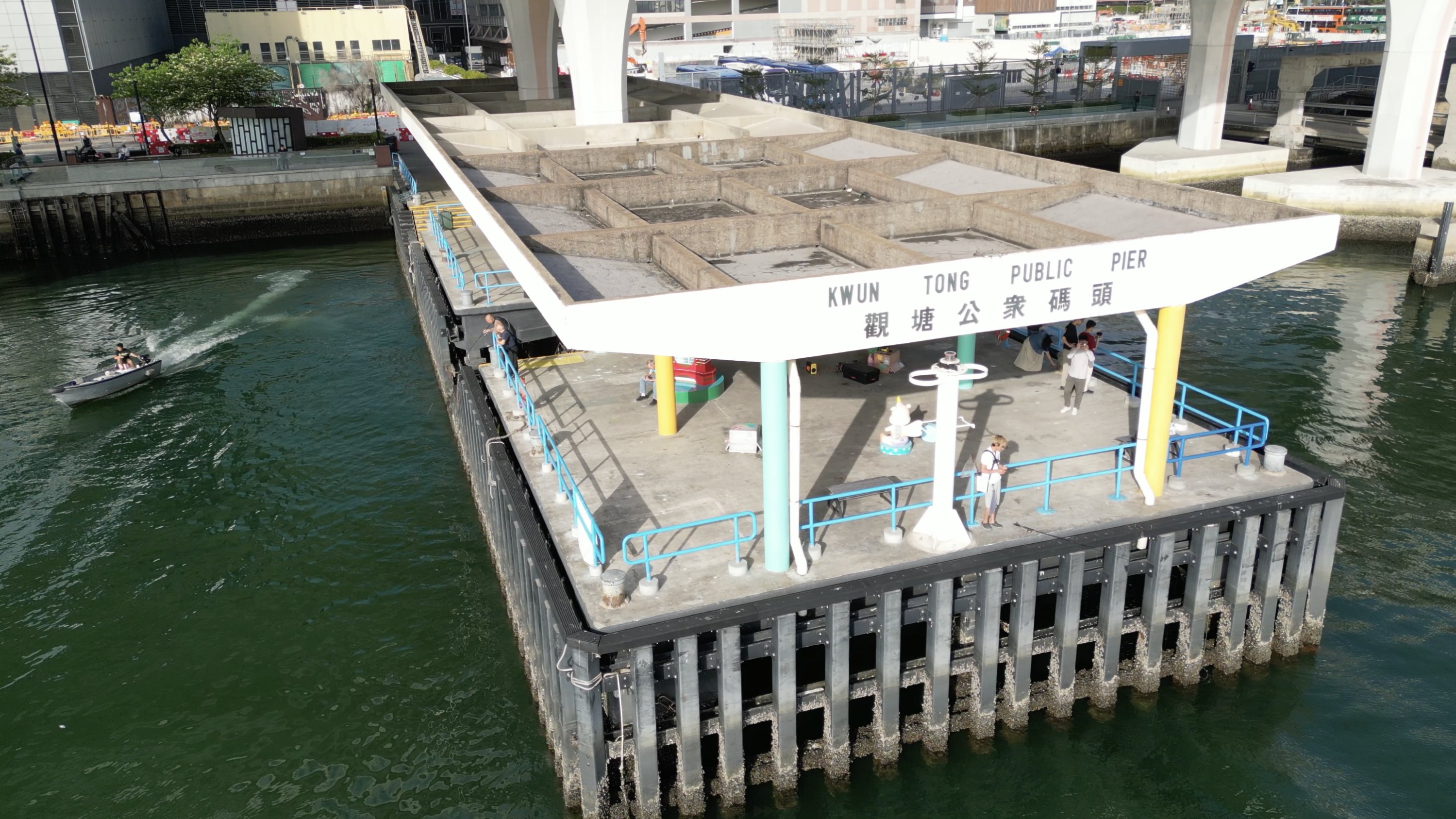}
        \caption{Kwun Tong Pier.}
        \label{fig:guan1}
    \end{subfigure}
    \begin{subfigure}{0.492\columnwidth}
        \includegraphics[width=\textwidth]{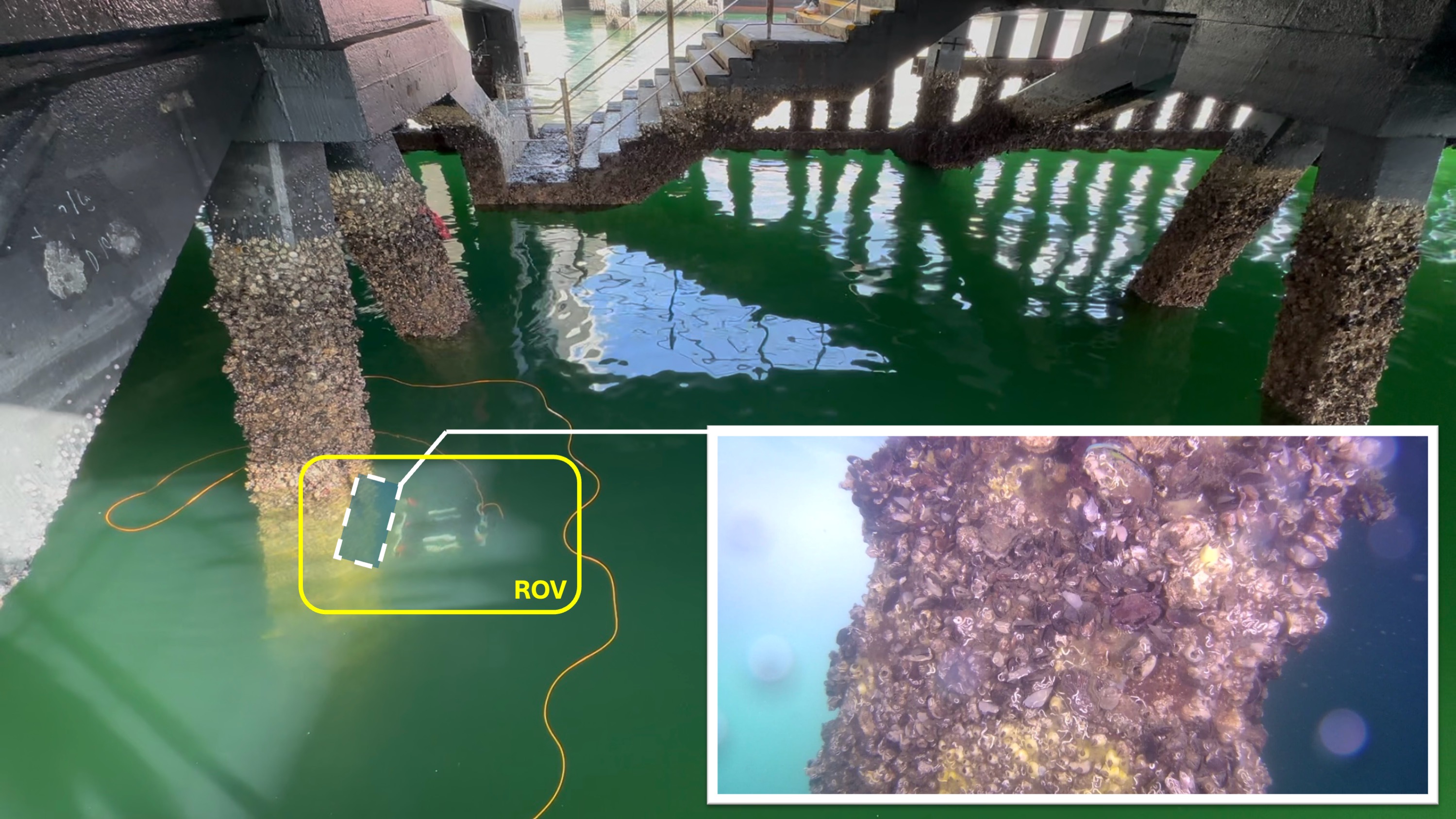}
        \caption{Underwater inspection.}
        \label{fig:guan2}
    \end{subfigure}
    \caption{\textcolor{blue}{Real ocean experiment 2 in Hong Kong.}}
    \label{fig:guantang}
\end{figure}

\textcolor{blue}{We also select another pier, the Kwun Tong Public Pier, as the second experimental scenario, as illustrated in Fig.~\ref{fig:guantang}. Compared to the Ma Liu Shui Pier, the Kwun Tong Pier has a similar overall structure, as shown in Fig.~\ref{fig:guan1}, indicating that our method can generalize well. Meanwhile, the Kwun Tong Pier features a more complex and cluttered internal structure, with an irregular hollow base and a combination of square and round columns. Additionally, we observe that the pier itself exhibits more severe corrosion, as illustrated in Fig.~\ref{fig:guan2}.}

\begin{figure}[htb]
    \centering
    \begin{subfigure}[b]{0.49\textwidth}
        \includegraphics[width=\textwidth]{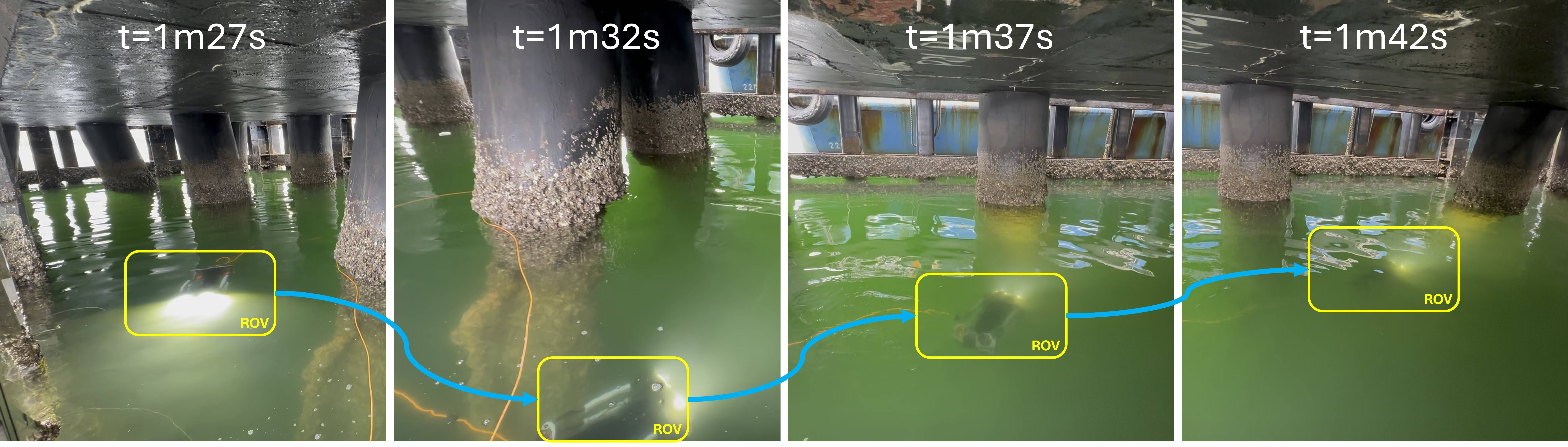} 
        \caption{The underwater robot executes along the planned path.}
        \label{fig:workflow1_1}
    \end{subfigure}
    \begin{subfigure}[b]{0.49\textwidth}
        \includegraphics[width=\textwidth]{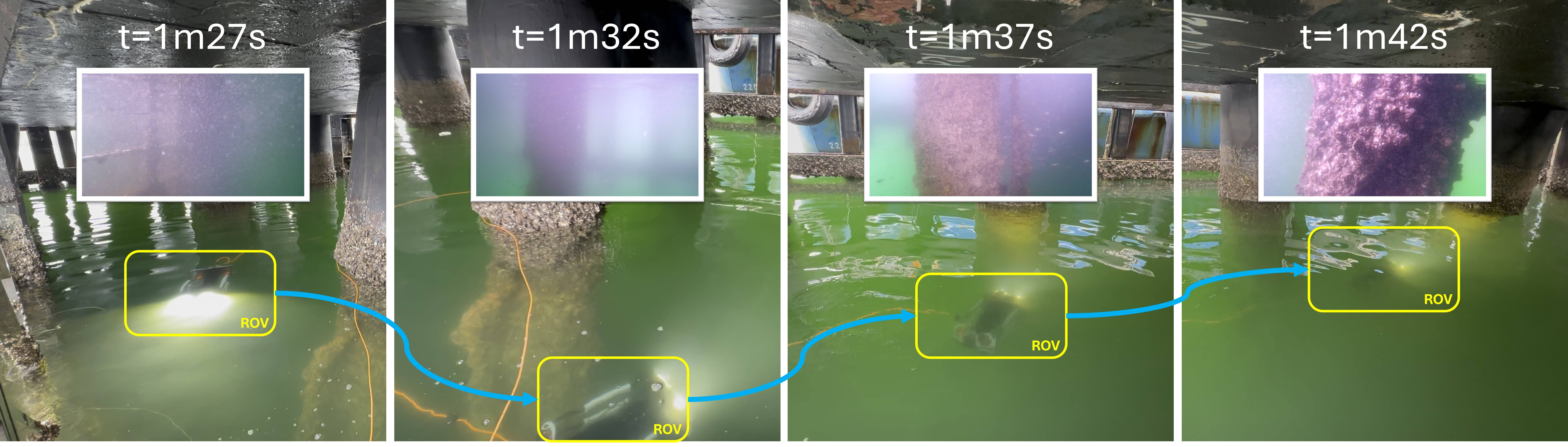} 
        \caption{Real-time observation is conducted during the process.}
        \label{fig:workflow1_2}
    \end{subfigure}
    \caption{Workflow of experiment 1 under Ma Liu Shui Pier.}
    \label{fig:workflow1}
\end{figure}

Fig.~\ref{fig:workflow1} further illustrates the workflow of the underwater robot's inspection experiment at the Ma Liu Shui Pier. In Fig.~\ref{fig:workflow1_1}, the robot moves to different positions along the pre-planned path at various times, with the blue arrows indicating the connecting lines between these positions. Fig.~\ref{fig:workflow1_2} shows the real-time observation images captured by the robot at selected moments. These images can be subsequently processed by the object detection system to identify whether there are any potential safety hazards.

\begin{figure}[htb]
    \centering
    \includegraphics[width=0.98\columnwidth]{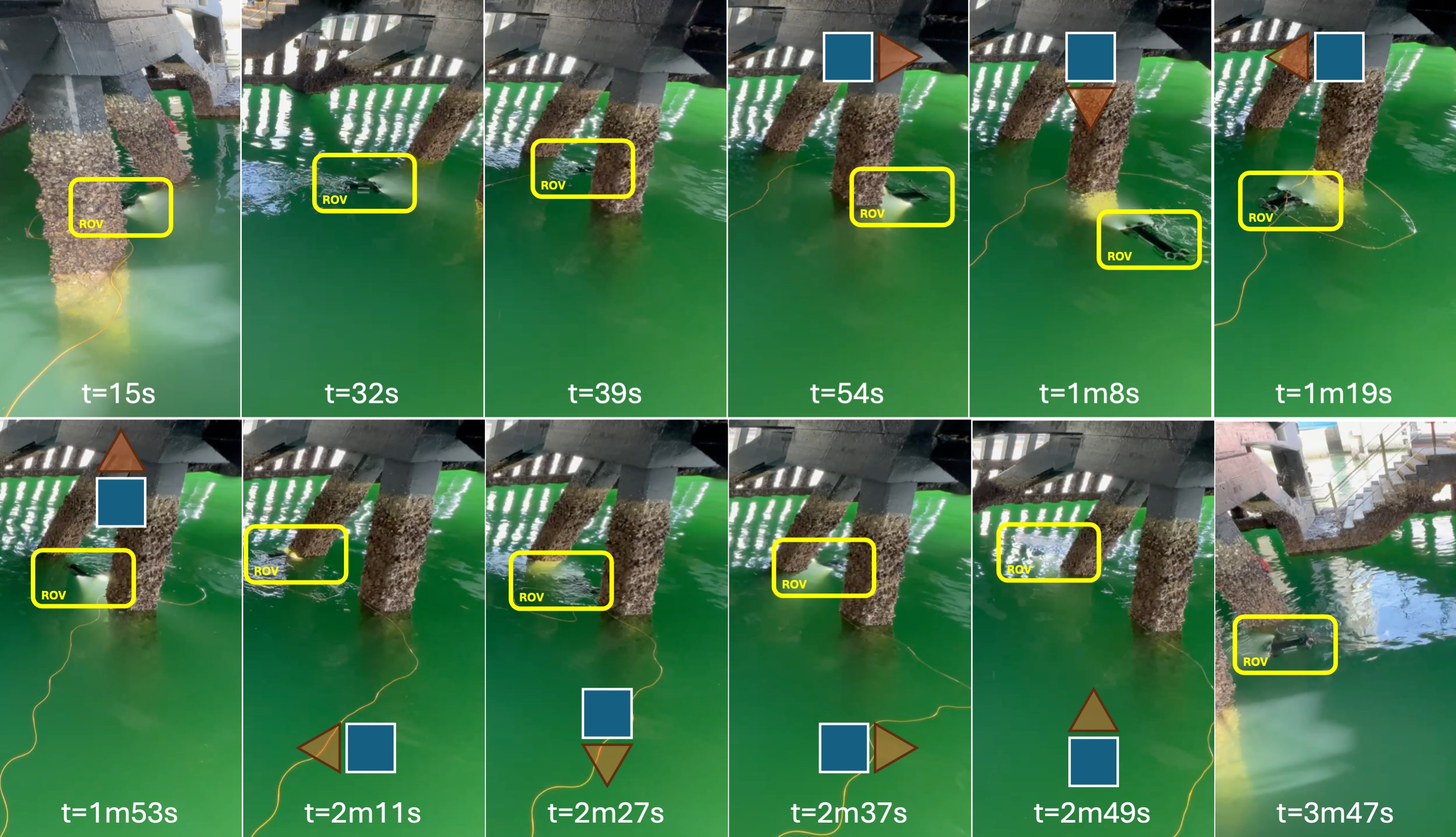} 
    \caption{\textcolor{blue}{Workflow of experiment 2 under Kwun Tong Pier.}}
    \label{fig:workflow2}
\end{figure}

\begin{figure}[htb]
    \centering
    \includegraphics[width=0.98\columnwidth]{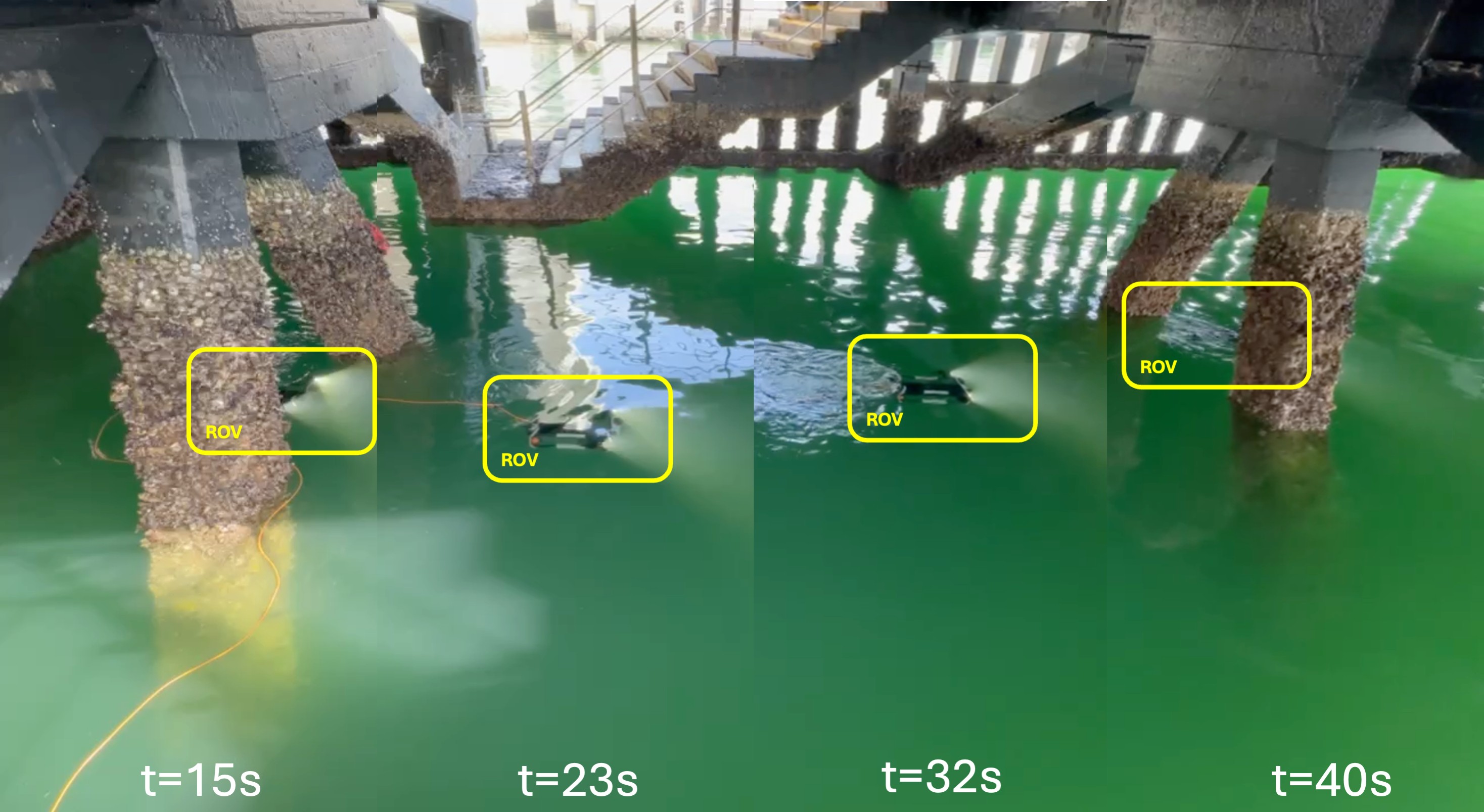} 
    \caption{\textcolor{blue}{Workflow of experiment 3 under Kwun Tong Pier.}}
    \label{fig:workflow3}
\end{figure}

\textcolor{blue}{Fig.~\ref{fig:workflow2} shows our testing scenario at the Kwun Tong Pier, which is more cluttered compared to the experiments conducted at the Ma Liu Shui Pier. We aim to guide the underwater robot to perform a "figure-eight" motion to enable comprehensive inspection of the underwater structures of the pier. Fig.~\ref{fig:workflow3} shows another test conducted at the Kwun Tong Pier. Compared to the previous field test, this scenario required the underwater robot to perform inspections over a larger scale. As can be seen, the robot navigates between two sets of pillars under the pier.}

\subsubsection{Sensor Data Visualization}

\begin{figure}[htb]
    \centering
    \begin{subfigure}{0.493\columnwidth}
        \includegraphics[width=\textwidth]{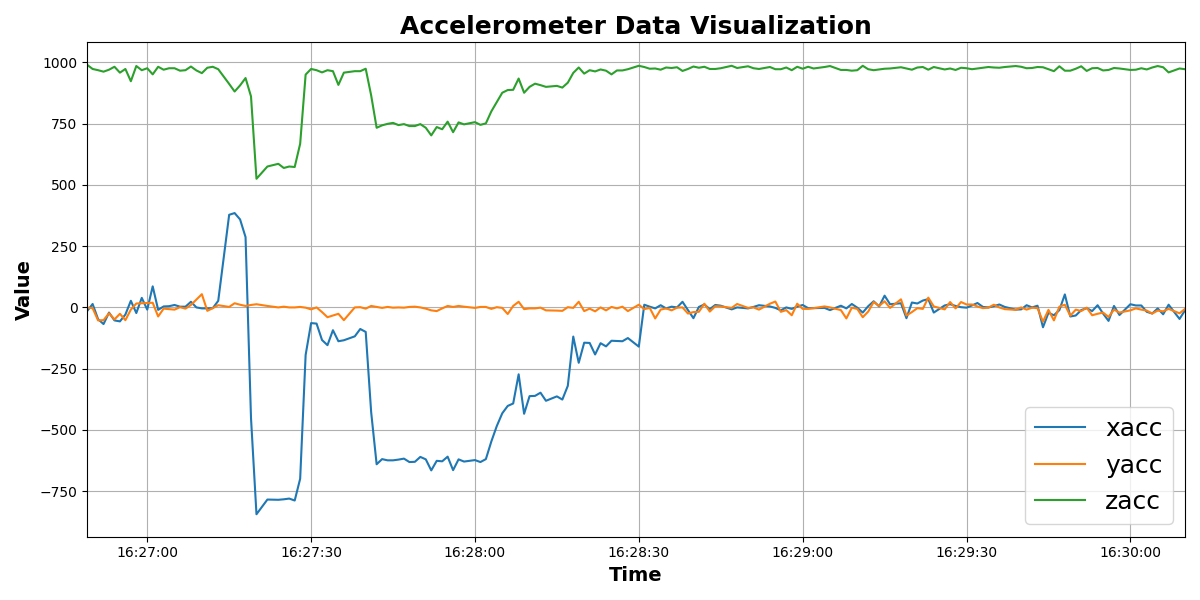}
        \caption{Accelerometer data 1.}
        \label{fig:acc_1}
    \end{subfigure}
    \begin{subfigure}{0.493\columnwidth}
        \includegraphics[width=\textwidth]{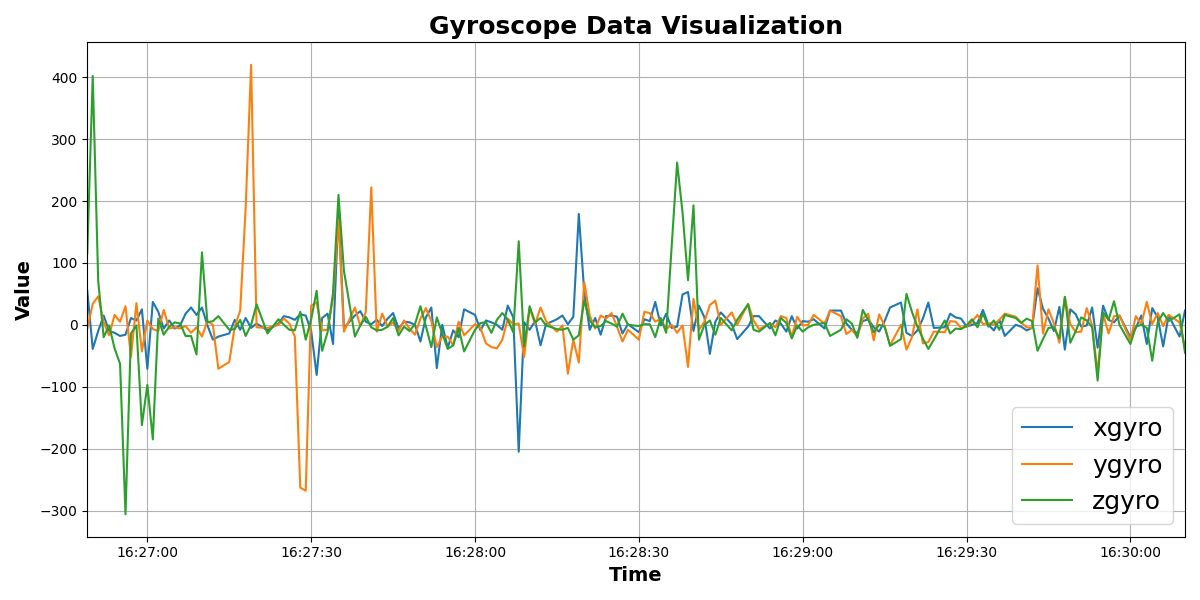}
        \caption{Gyroscope data 1.}
        \label{fig:gyro_1}
    \end{subfigure}
    \caption{\textcolor{blue}{Visualization of sensor data in experiment 1.}}
    \label{fig:data1}
\end{figure}

\begin{figure}[htb]
    \centering
    \begin{subfigure}{0.493\columnwidth}
        \includegraphics[width=\textwidth]{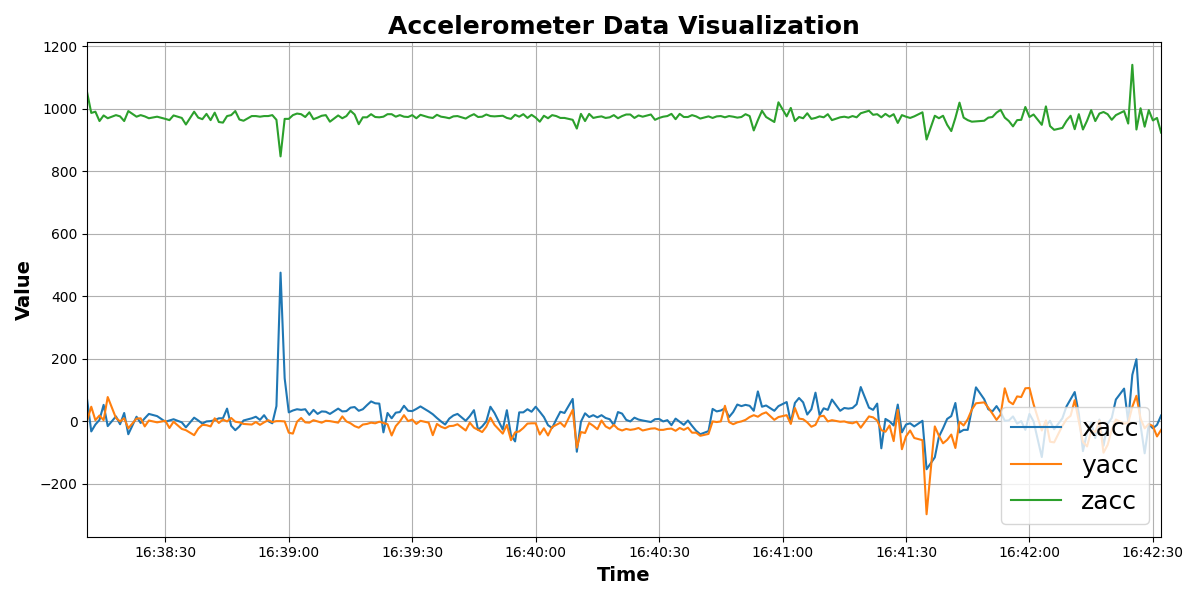}
        \caption{Accelerometer data 2.}
        \label{fig:acc_2}
    \end{subfigure}
    \begin{subfigure}{0.493\columnwidth}
        \includegraphics[width=\textwidth]{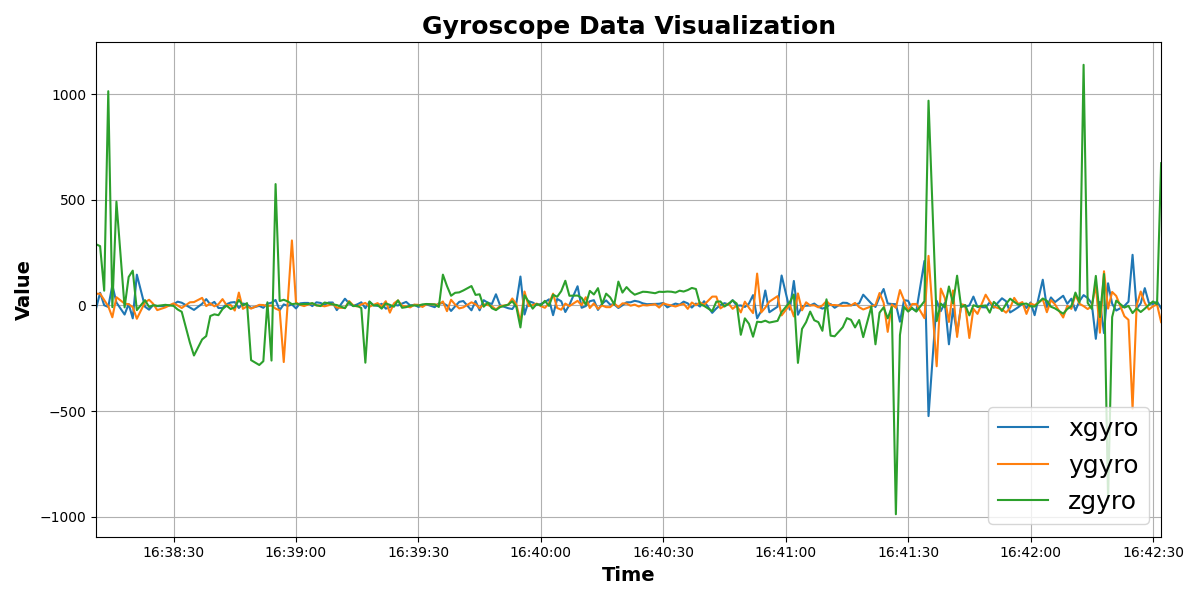}
        \caption{Gyroscope data 2.}
        \label{fig:gyro_2}
    \end{subfigure}
    \caption{\textcolor{blue}{Visualization of sensor data in experiment 2.}}
    \label{fig:data2}
\end{figure}

\textcolor{blue}{Our underwater robotic platform is equipped with multiple onboard sensors, including a depth sensor, gyroscope, accelerometer, and magnetometer, all of which require calibration prior to deployment. To analyze the motion dynamics and attitude variation of the underwater robot during inspection tasks, we collected and visualized six-axis inertial data, including three-axis accelerometer and three-axis gyroscope readings. Two inspection episodes were selected, and their corresponding time-series sensor signals are shown in Fig.~\ref{fig:data1} - Fig.~\ref{fig:data2}. The gyroscope data (xgyro, ygyro, zgyro) effectively captures rotational changes, highlighting angular disturbances and maneuvers throughout the mission, as shown in Fig.~\ref{fig:gyro_1} - Fig.~\ref{fig:gyro_2}. Meanwhile, the accelerometer data (xacc, yacc, zacc) reflects both gravitational orientation and transient linear accelerations, allowing us to identify significant transitions in vertical and horizontal movements, as shown in Fig.~\ref{fig:acc_1} - Fig.~\ref{fig:acc_2}. Notably, variations in z-axis acceleration (zacc) are particularly indicative of depth change dynamics or abrupt pitch adjustments.} 

\textcolor{blue}{Additionally, the underwater experiments were conducted beneath a pier, where the supporting structures naturally mitigated strong water currents, resulting in relatively calm conditions. Although underwater visibility can be a challenge, the robot is equipped with a high-resolution camera supporting 4K video recording and 12-megapixel image capture, which is sufficient for typical pier inspection tasks. Communication and power were ensured through a tether connecting the robot to the surface, allowing for stable real-time operation throughout the missions.}

\subsubsection{Safety Protocol}

\textcolor{blue}{In our current framework, safety during path planning is primarily ensured by strict collision checking based on the known environment map, which guarantees that all generated paths are free of obstacle collisions. Additionally, we include validity checks for sampled nodes and paths at each iteration, ensuring that the planner does not proceed with unsafe connections. In real deployments, we integrate additional safety mechanisms to ensure system-level robustness and mission continuity. These include automatic emergency surfacing triggered upon detecting abnormal sensor readings—such as communication loss, severe attitude instability, or excessive deviation from the planned path—as well as physical constraint enforcement that rejects any control commands exceeding the robot’s mechanical limits.}

\subsection{Result Analysis}

\subsubsection{Quantitative Comparison}

\begin{figure*}[htb]
    \centering
    \begin{subfigure}{0.24\textwidth}
        \includegraphics[width=\textwidth]{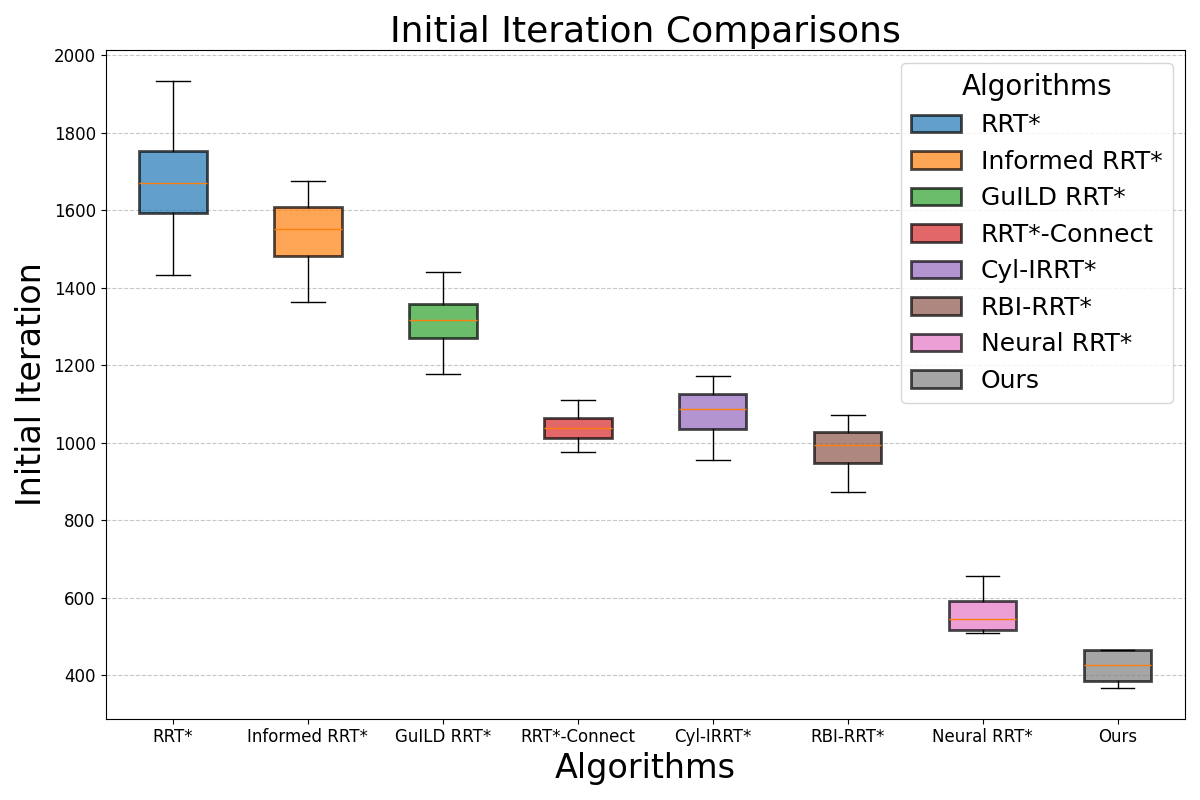}
        \caption{Initial Iteration.}
        \label{fig:simu1_1}
    \end{subfigure}
    \begin{subfigure}{0.24\textwidth}
        \includegraphics[width=\textwidth]{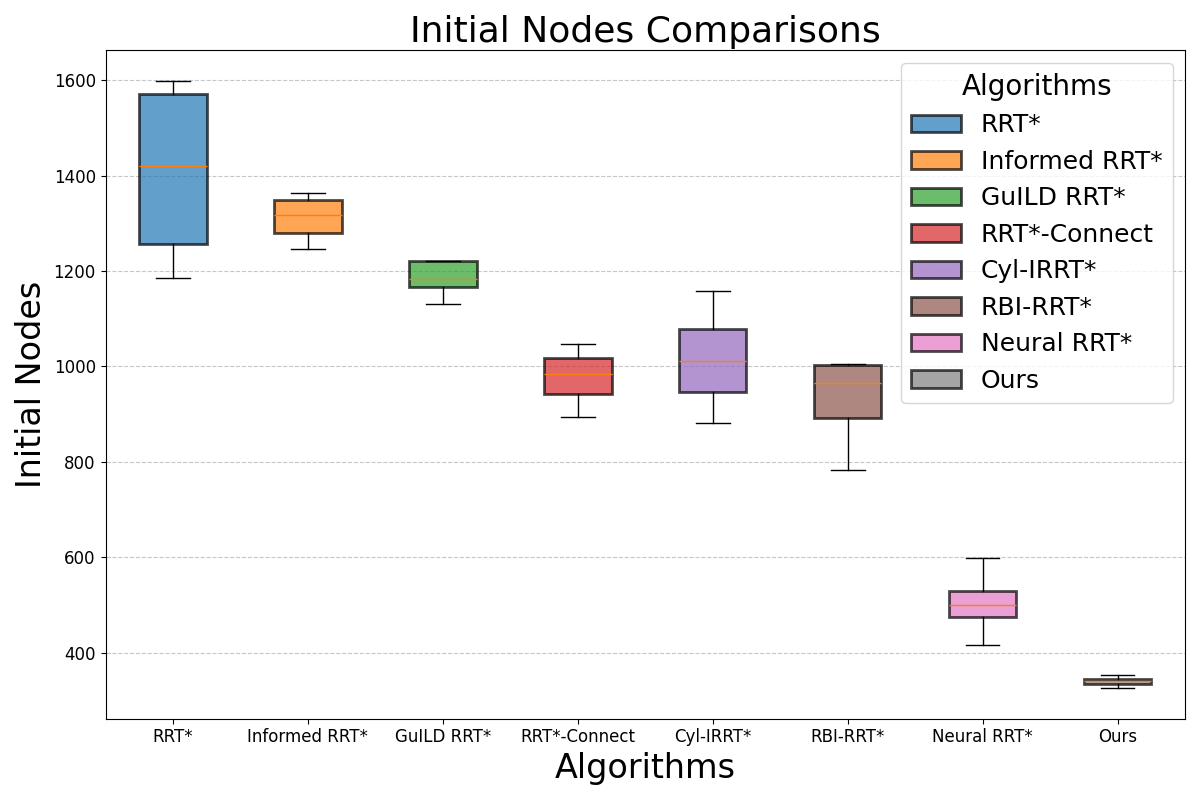}
        \caption{Initial Nodes.}
        \label{fig:simu1_2}
    \end{subfigure}
    \begin{subfigure}{0.24\textwidth}
        \includegraphics[width=\textwidth]{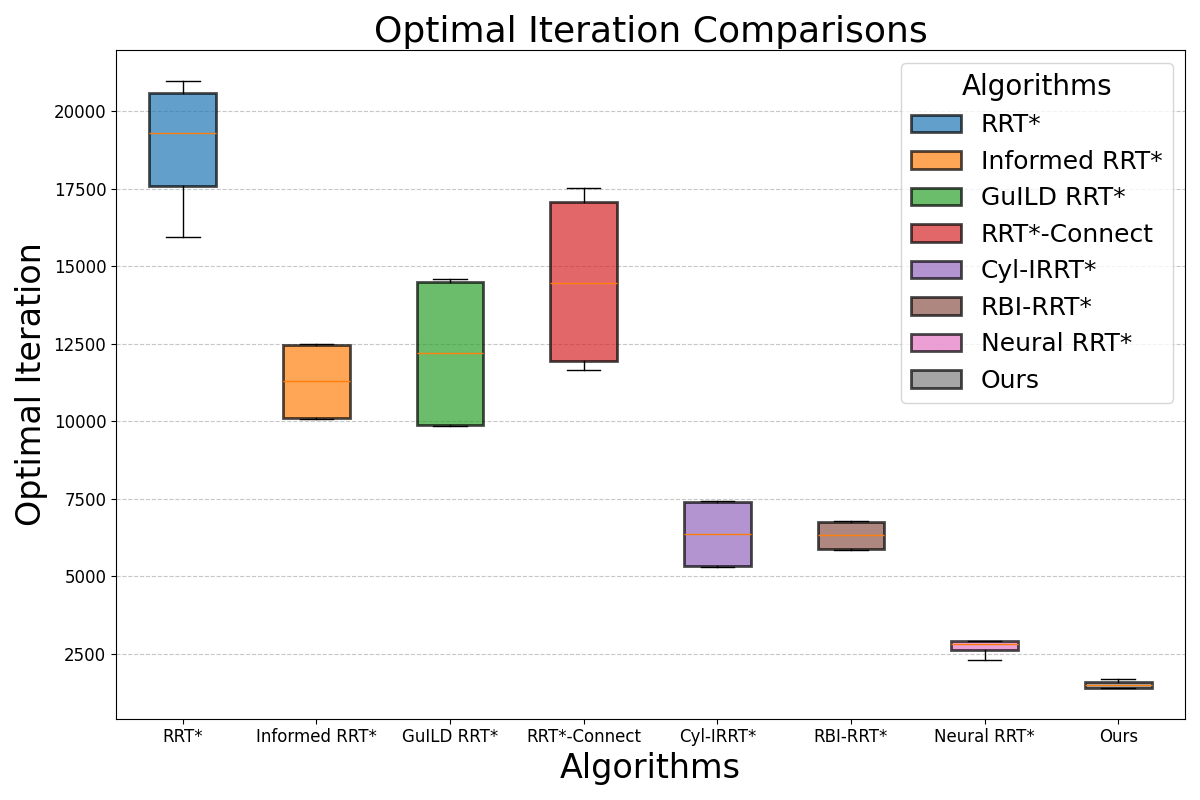}
        \caption{Optimal Iteration.}
        \label{fig:simu1_3}
    \end{subfigure}
    \begin{subfigure}{0.24\textwidth}
        \includegraphics[width=\textwidth]{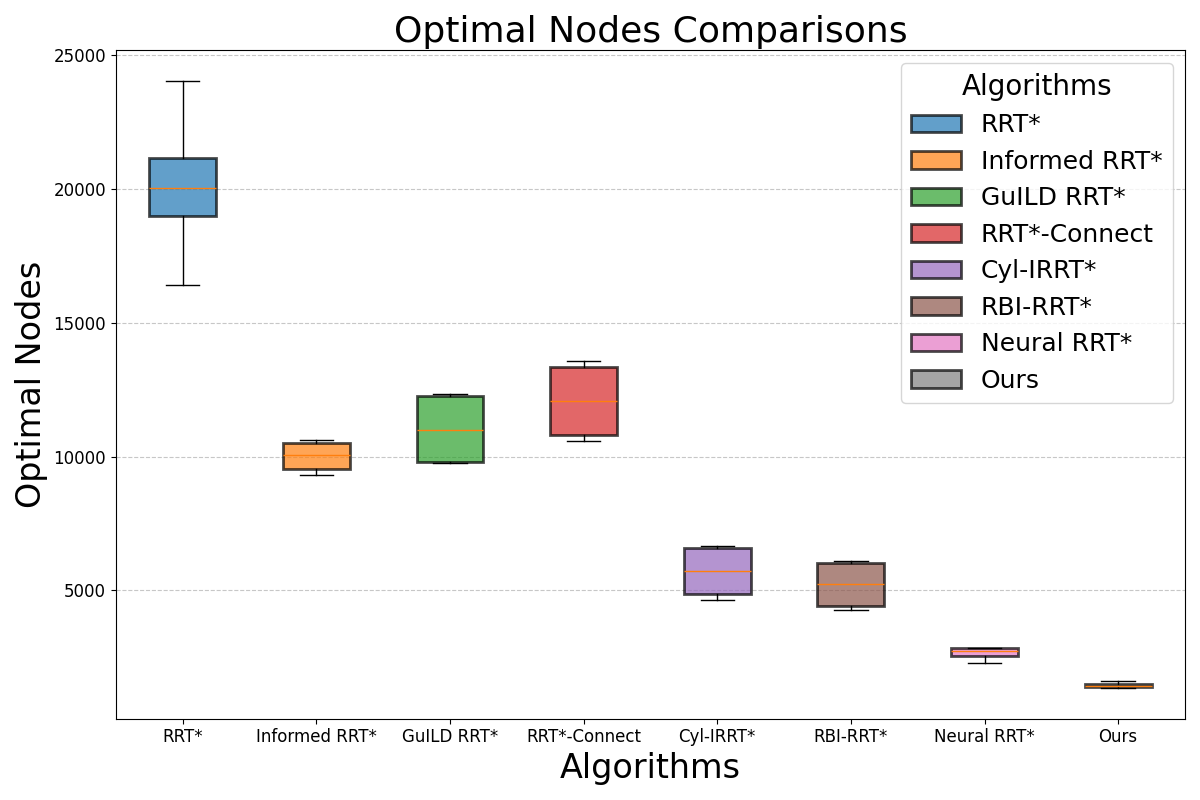}
        \caption{Optimal Nodes.}
        \label{fig:simu1_4}
    \end{subfigure}
    \caption{Comparisons in terms of iteration and nodes in the initial phase and optimal phase on map 1 (For each subfigure, from left to right: RRT*, Informed-RRT*, GuILD-RRT*, RRT*-Connect, Cyl-IRRT*, RBI-RRT*, Neural RRT*, Ours).}
    \label{fig:Map1}
\end{figure*}

\begin{figure*}[htb]
    \centering
    \begin{subfigure}{0.24\textwidth}
        \includegraphics[width=\textwidth]{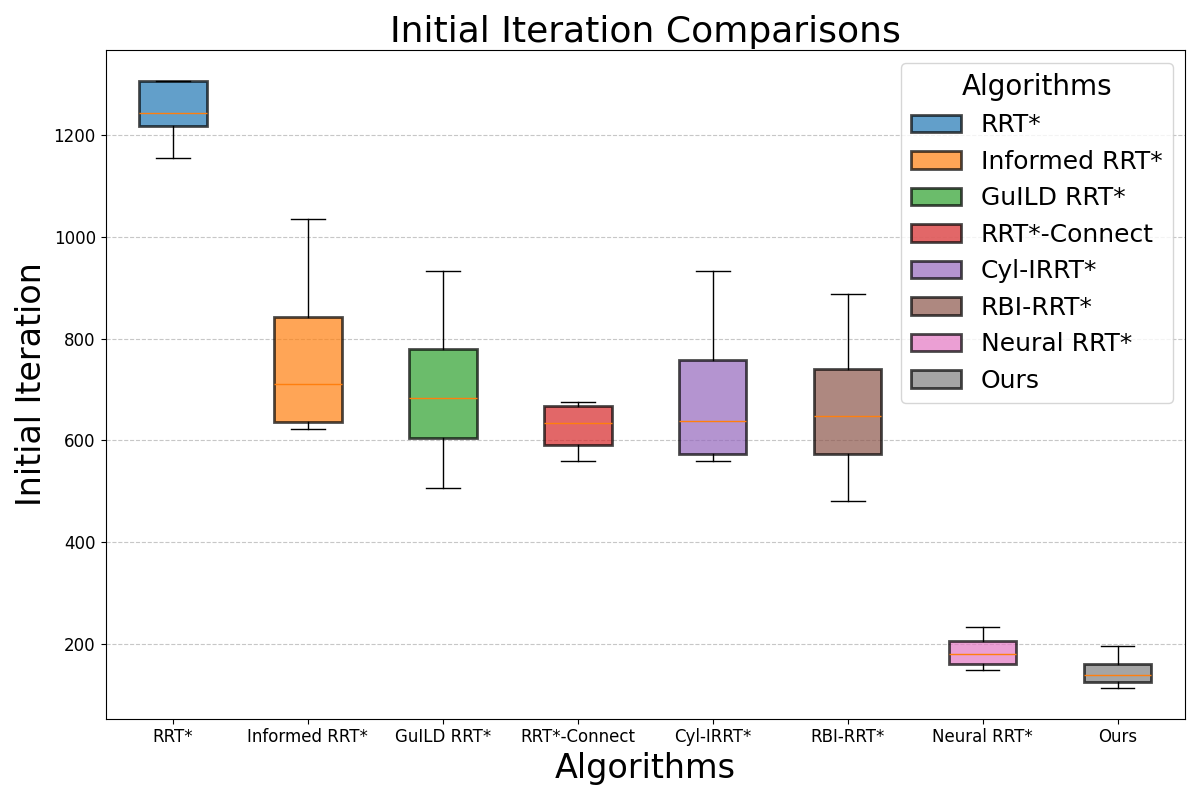}
        \caption{Initial Iteration.}
        \label{fig:simu2_1}
    \end{subfigure}
    \begin{subfigure}{0.24\textwidth}
        \includegraphics[width=\textwidth]{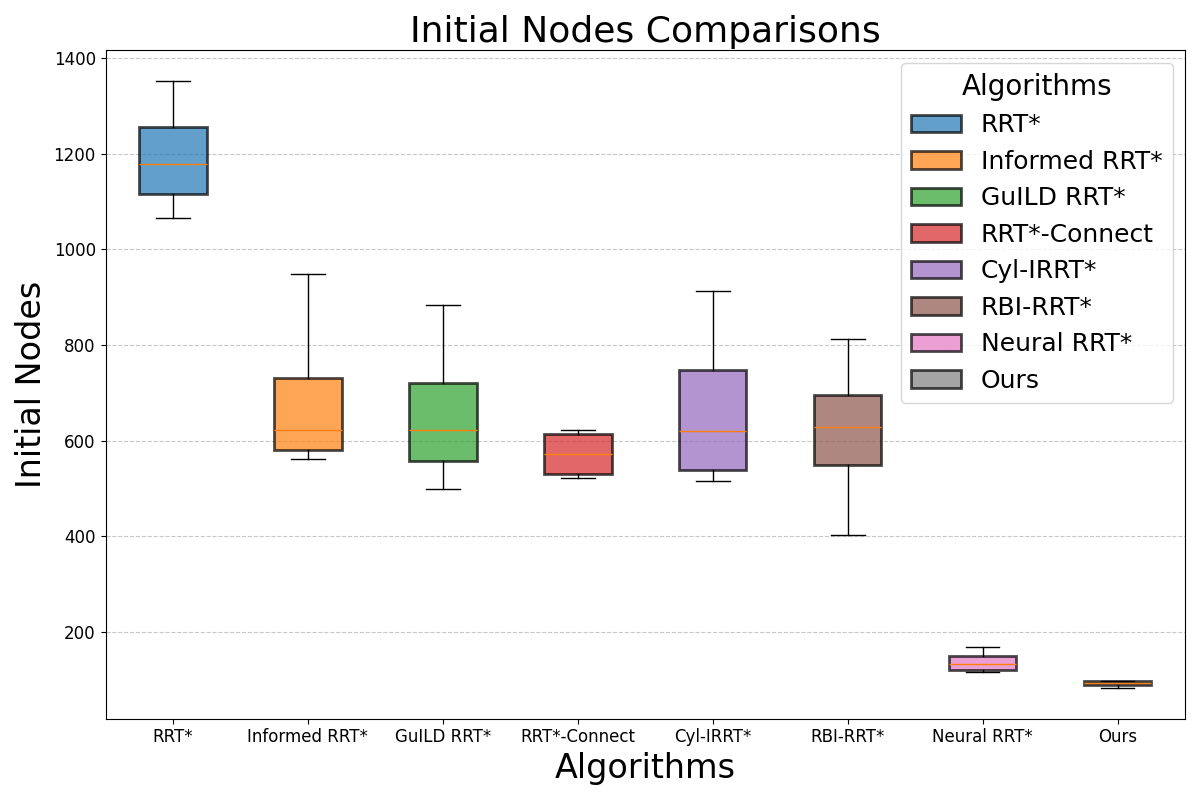}
        \caption{Initial Nodes.}
        \label{fig:simu2_2}
    \end{subfigure}
    \begin{subfigure}{0.24\textwidth}
        \includegraphics[width=\textwidth]{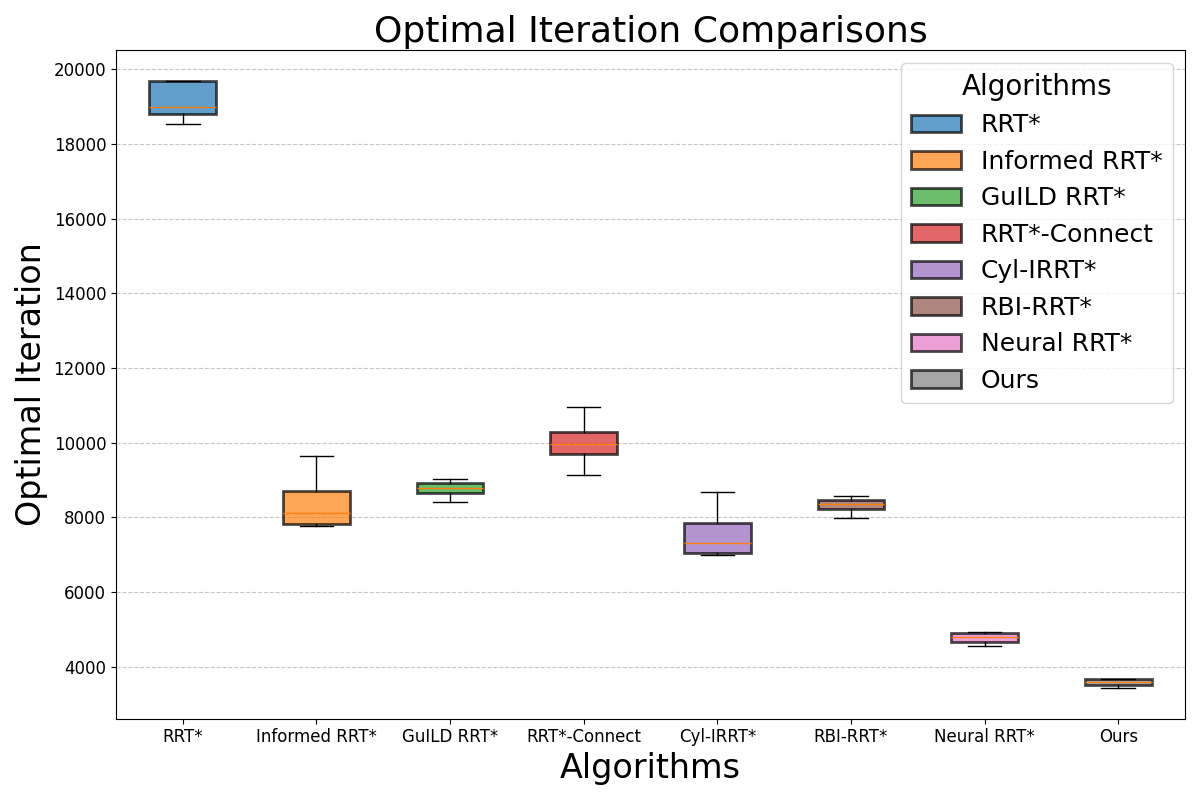}
        \caption{Optimal Iteration.}
        \label{fig:simu2_3}
    \end{subfigure}
    \begin{subfigure}{0.24\textwidth}
        \includegraphics[width=\textwidth]{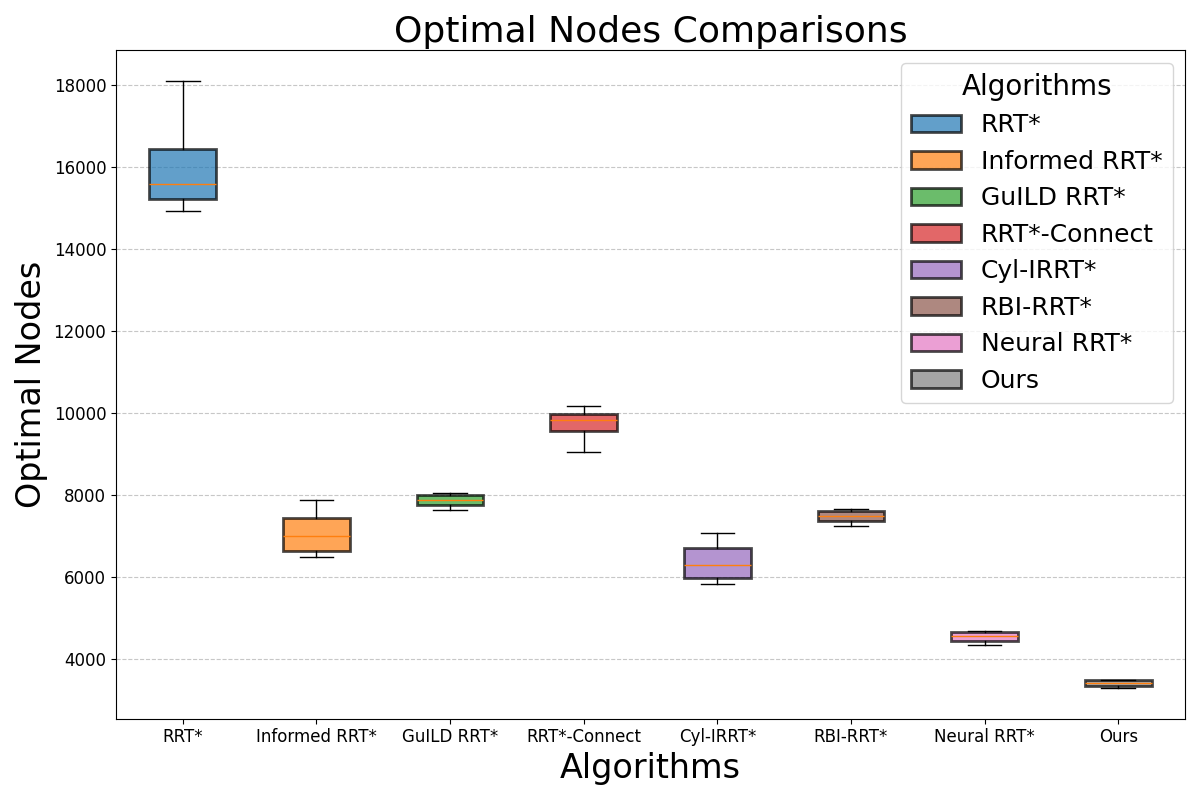}
        \caption{Optimal Nodes.}
        \label{fig:simu2_4}
    \end{subfigure}
    \caption{Comparisons in terms of iteration and nodes in the initial phase and optimal phase on map 2 (For each subfigure, from left to right: RRT*, Informed-RRT*, GuILD-RRT*, RRT*-Connect, Cyl-IRRT*, RBI-RRT*, Neural RRT*, Ours).}
    \label{fig:Map2}
\end{figure*}

\textcolor{blue}{We conduct extensive comparative experiments and quantitative data analysis between our method and classical baseline approaches~\cite{karaman2011sampling, gammell2018informed, kuffner2000rrt, JordanPerez2013}, as well as the state-of-the-art geometry-heuristic methods~\cite{scalise2023guild, yu2023cyl, chen2024rbi, ayalew2024optimal} and neural-heuristic methods~\cite{wang2020neural, wang2021deep}.} We employ seven quantitative metrics, applying four in the initial phase and three in the optimal phase. Among these metrics, iteration represents the number of main program cycles, nodes indicate the count of newly generated tree nodes, cost denotes the path length measured by Euclidean distance, and time refers to the execution duration of the main program, as shown in Tab.~\ref{tab:table1} and Fig.~\ref{fig:Map1} - Fig.~\ref{fig:Map2}.

\begin{table*}[t]
\caption{Comparison of Different Heuristic-based RRT* Algorithms on Two Representative Maps\label{tab:table1}}
\centering
\renewcommand{\arraystretch}{1.15}
\begin{tabular}{c | c | c c c c c c c}
\hline
\multirow{2}{*}{Map} & \multirow{2}{*}{Method} & \multirowcell{2}{Initial Iteration} & \multirowcell{2}{Initial Nodes} & \multirowcell{2}{Initial Time} & \multirowcell{2}{Initial Cost} & \multirowcell{2}{Optimal Iteration} & \multirowcell{2}{Optimal Nodes} & \multirowcell{2}{Optimal Time}\\
& & & & & & & \\
\hline
\multirow{9}{*}{1} & RRT*~\cite{karaman2011sampling} & 1683.6 & 1372 & 4.28 & 107.16 & 21059.0 & 20199.4 & 200.61 \\
                   & Informed-RRT*~\cite{gammell2018informed} & 1594.2 & 1292.6 & 4.03 & 107.91 & 11031.6 & 10005.8 & 99.36\\
                   & GuILD-RRT*~\cite{scalise2023guild} & 1295.2 & 1201.8 & 3.75 & 103.90 & 12782.0 & 11398.0 & 113.18\\
                   & RRT*-connect~\cite{kuffner2000rrt, JordanPerez2013} & 1036.4 & 976.8 & 3.05 & 103.78 & 14440.2 & 12172.4 & 120.87\\
                   & \textcolor{blue}{BRRT*-DWA}~\cite{ayalew2024optimal} & 1042.1 & 982.5 & 3.08 & 103.82 & 14235.8 & 12045.3 & 119.15 \\
                   & Cyl-IRRT*~\cite{yu2023cyl} & 1115.9 & 1061.1 & 3.31 & 106.45 & 6877.6 & 6055.1 & 70.12\\
                   & RBI-RRT*~\cite{chen2024rbi} & 1020.2 & 972.9 & 3.03 & 109.10 & 6505.3 & 5944.6 & 69.02\\
                   & Neural RRT*~\cite{wang2020neural,wang2021deep} & 588.4 & 533.2 & 1.66 & 93.90 & 2728.0 & 2668.4 & 26.50\\
                   & \textbf{Ours} & \textbf{430.8} & \textbf{383.4} & \textbf{1.20} & \textbf{92.10} & \textbf{1584.4} & \textbf{1412.8} & \textbf{14.03}\\      
\hline
\multirow{9}{*}{2} & RRT*~\cite{karaman2011sampling} & 1293.8 & 1186.8 & 3.71 & 107.69 & 19771.0 & 16350.4 & 189.14\\
                   & Informed-RRT*~\cite{gammell2018informed} & 839.6 & 768.6 & 2.40 & 107.99 & 8580.0 & 7177.4 & 83.05\\
                   & GuILD-RRT*~\cite{scalise2023guild} & 657.2 & 632.4 & 1.98 & 107.50 & 9062.2 & 7778.8 & 90.00\\
                   & RRT*-connect~\cite{kuffner2000rrt, JordanPerez2013} & 630.4 & 607.2 & 1.89 & 109.33 & 10168.6 & 9526.2 & 110.20\\
                   & \textcolor{blue}{BRRT*-DWA}~\cite{ayalew2024optimal} & 625.8 & 602.4 & 1.87 & 108.95 & 10142.3 & 9480.7 & 109.85 \\
                   & Cyl-IRRT*~\cite{yu2023cyl} & 755.6 & 721.7 & 2.26 & 102.35 & 7722.0 & 6459.7 & 79.74\\
                   & RBI-RRT*~\cite{chen2024rbi} & 624.3 & 604.8 & 1.88 & 108.30 & 8230.1 & 7389.9 & 85.51\\
                   & Neural RRT*~\cite{wang2020neural,wang2021deep} & 191.4 & 138.4 & 0.43 & 91.11 & 4714.0 & 4494.4 & 42.00\\
                   & \textbf{Ours} & \textbf{153.2} & \textbf{107.0} & \textbf{0.33} & \textbf{90.40} & \textbf{3569.4} & \textbf{3406.2} & \textbf{29.41}\\      
\hline
\end{tabular}
\end{table*}

As can be observed from Tab.~\ref{tab:table1}, compared to existing methods, our approach demonstrates superior performance in the initial phase by identifying high-quality initial solutions with lower costs, while requiring fewer iterations, fewer nodes, and less computational time. Leveraging the high-quality initial solutions and the inherent advantages of our method, our approach also achieves the fastest convergence in the final stage of finding the asymptotically optimal path. Our method outperforms Neural RRT* by specifically addressing underwater cluttered environments, yielding heuristic regions with fewer obstacles and improved connectivity. Additionally, our bidirectional search improves exploration efficiency, further enhancing performance. For each fixed map, the optimal cost is predetermined and remains consistent across all methods. Therefore, reaching this optimal cost is considered as finding the optimal solution, which is why it has not been explicitly annotated in the table. \textcolor{blue}{Overall, from the perspective of actual execution, our method demonstrates the lowest space overhead (as indicated by the Iteration and Nodes metrics) and the lowest time overhead (as indicated by the Time metric).}

Additionally, we have visualized the mean and variance of all methods using boxplots. As shown in Fig.~\ref{fig:Map1} - Fig.~\ref{fig:Map2}, our method demonstrates nearly the smallest variance, indicating superior performance in both individual runs and average performance metrics. Our methodology demonstrates transferability, as port facilities worldwide share common characteristics of having fixed yet complex underwater structural configurations following their construction. The inherent efficiency and robustness of our approach enable its adaptation to planning and inspection tasks across virtually any port infrastructure globally, thereby offering practical insights.

By demonstrating an autonomous and efficient underwater inspection approach, our work offers insights for enhancing maritime facility resilience. The developed methodologies are adaptable to diverse environments, advancing intelligent maintenance and ensuring global shipping safety.

\subsubsection{Ablation Study}

\textcolor{blue}{To analyze the contribution of each component in our framework, we conduct an ablation study comparing four loss functions: the standard Binary Cross-Entropy $\mathcal{L}_{\text{BCE}}$, our path-weighted variant $\mathcal{L}_{\text{path}}$ that emphasizes critical regions near the ground-truth path, our Hausdorff distance loss $\mathcal{L}_{\text{haus}}$ for topological consistency, and their combined form $\mathcal{L}_{\text{total}}$. As shown in Tab.~\ref{tab:ablation}, we evaluate three key metrics: Connectivity measures the success rate of finding progressive optimal paths when sampling exclusively within the predicted heuristic region. Initial Iteration and Optimal Iteration record the required planning iterations on Map 1 as an example to obtain initial feasible paths and converged optimal solutions, respectively. This systematic comparison demonstrates how each loss component contributes to the overall planning performance.}

\begin{table}[h]
\centering
\caption{\textcolor{blue}{Ablation study of different loss components}}
\label{tab:ablation}
\renewcommand{\arraystretch}{1.5}
\begin{tabular}{c|c c c}
\hline
Method & Connectivity & Initial Iteration & Optimal Iteration \\
\hline
$\mathcal{L}_{\text{BCE}}$   & 90.4\%       & 612.3       & 2250.6       \\
$\mathcal{L}_{\text{path}}$  & 95.7\%       & 518.2       & 1903.9       \\
$\mathcal{L}_{\text{haus}}$  & 93.8\%       & 554.7       & 2038.2       \\
$\mathcal{L}_{\text{total}}$ & \textbf{97.8\%}      & \textbf{430.8}      & \textbf{1584.4}       \\
\hline
\end{tabular}
\end{table}

\textcolor{blue}{To better demonstrate the scalability and robustness of our method, we conduct tests on various map sizes (32×32x32, 64×64x64, 128×128x128) and obstacle densities (10\% and 30\%). As shown in Tab.~\ref{tab:scalability}, our method achieves strong performance across different map settings, which also verifies the generalizability of our approach.}

\begin{table}[ht]
\centering
\caption{\textcolor{blue}{Planning performance across different map sizes and obstacle densities}}
\renewcommand{\arraystretch}{1.5}
\begin{tabular}{c|c c c c c}
\hline
\makecell{Map\\Size} & 
\makecell{Obstacle\\Density} & 
\makecell{Initial\\Iteration} & 
\makecell{Initial\\Nodes} & 
\makecell{Optimal\\Iteration} & 
\makecell{Optimal\\Nodes} \\
\hline
32×32×32 & 10\% & 180.4 & 120.2 & 1350.6 & 850.3 \\
32×32×32 & 30\% & 220.8 & 150.5 & 1680.3 & 1050.7 \\
\hline
64×64×64 & 10\% & 300.2 & 200.1 & 2320.5 & 1450.8 \\
64×64×64 & 30\% & 360.5 & 240.3 & 2790.2 & 1750.4 \\
\hline
128×128×128 & 10\% & 520.7 & 350.4 & 3560.3 & 2200.6 \\
128×128×128 & 30\% & 680.6 & 450.2 & 4350.1 & 2700.3 \\
\hline
\end{tabular}
\label{tab:scalability}
\end{table}

\section{CONCLUSION}
This paper presents PierGuard, a fast and efficient global path planning framework for underwater inspection of pier structures. By integrating the strengths of bidirectional search and a novel neural network that generates high-quality heuristic regions, the framework achieves superior planning performance in cluttered underwater environments. Its effectiveness is validated through extensive simulations and real-world ocean experiments in Hong Kong, highlighting its potential for practical applications in challenging marine pier settings. In the future, we aim to achieve full automation of underwater robot autonomous planning and detection systems, enabling their deployment for various inspection tasks at piers in coastal cities.

\bibliographystyle{ieeetr}
\bibliography{main}

\end{document}